\documentclass[aps,prl,onecolumn,notitlepage,nofootinbib,superscriptaddress]{revtex4-1}
\usepackage{amssymb,amsfonts,amsmath}
\usepackage{euscript}
\usepackage{graphicx}
\usepackage{xcolor}

\begin{document}

\title{Optimal modularity and memory capacity of neural reservoirs}

\author{Nathaniel Rodriguez}
\affiliation{School of Informatics and Computing, Indiana University, Bloomington, Indiana, USA}
\author{Eduardo Izquierdo}
\affiliation{Cognitive Science Program, Indiana University, Bloomington, Indiana, USA}
\author{Yong-Yeol Ahn}
\email{yyahn@indiana.edu}
\affiliation{School of Informatics and Computing, Indiana University, Bloomington, Indiana, USA}
\affiliation{Indiana University Network Science Institute, Bloomington, Indiana, USA}

\begin{abstract}
The neural network is a powerful computing framework that has been exploited by biological evolution and by humans for solving diverse problems. Although the computational capabilities of neural networks are determined by their structure, the current understanding of the relationships between a neural network's architecture and function is still primitive. Here we reveal that neural network's modular architecture plays a vital role in determining the neural dynamics and memory performance of the network of threshold neurons. In particular, we demonstrate that there exists an optimal modularity for memory performance, where a balance between local cohesion and global connectivity is established, allowing optimally modular networks to remember longer. Our results suggest that insights from dynamical analysis of neural networks and information spreading processes can be leveraged to better design neural networks and may shed light on the brain's modular organization.
\end{abstract}

\maketitle

\section{Introduction}
Neural networks are the computing engines behind many living organisms. They are also prominent general-purpose frameworks for machine learning and artificial intelligence applications~\cite{LeCun2015}. The behavior of a neural network is determined by the dynamics of individual neurons, the topology and strength of individual connections, and large-scale architecture. Both in biological and artificial neural networks, neurons integrate input signals and produce a graded or threshold-like response. While individual connections are dynamically trained and adapted to the specific environment, the architecture primes the network for performing specific types of tasks. The architecture of neural networks vary from organism to organism and between brain regions and are vital for functionality. For instance, the orientation columns of the visual cortex that support low-level visual processing~\cite{Wiesel1972}, or the looped structure of hippocampus which consolidates memory~\cite{Baldassarre2013}. In machine learning, feed-forward convolutional architectures have achieved super-human visual recognition capabilities~\cite{LeCun2015, Ioffe2015}, while recurrent architectures exhibit impressive natural language processing and control capabilities~\cite{Schmidhuber2015}.

Yet, identifying systematic design principles for neural architecture is still an outstanding question~\cite{Legenstein2005,Sussillo2013}.
Here, we investigate the role of modular architectures on memory capacity of neural networks, where we define modules (communities) as groups of nodes that have stronger internal versus external connectivity~\cite{Girvan2002}.

We focus on modularity primarily because of the prevalence of modular architectures in the brain. 
Modularity can be observed across all scales in the brain and is considered a key organizing principle for functional division of brain regions~\cite{Bullmore2009} and brain dynamics~\cite{Muller-Linow2008, Kaiser2010, Wang2011, Moretti2013, Villegas2014}, and is also considered as a plausible mechanism for working memory through ensemble based coding schemes~\cite{Boerlin2011}, bistability~\cite{Constantinidis2016, Cossart2003, Klinshov2014}, gating~\cite{Gisiger2011}, and through metastable states that retain information~\cite{Johnson2013}.

Here we study the role of modularity based on the theories of information diffusion, which can inform how structural properties affect spreading processes on a network~\cite{Misic2015}. Spreading processes can include diseases, social fads, memes, random walks, or the spiking events transmitted by biological neurons~\cite{Newman2003a, Boccaletti2006, Pastor-Satorras2015}, and they are studied in the context of large-scale network properties like small-worldness, scale-freeness, core periphery structure, and community structure (modularity)~\cite{Newman2003a, Boccaletti2006, Strogatz2001}.

Communities' main role in information spreading is restricting information flow~\cite{Chung2014, Onnela2007}. However, recent work showed that communities may play a more nuanced role in \emph{complex contagions}, which require reinforcement from multiple local adoptions. It turns out that under certain conditions community structure can facilitate spread of complex contagions, mainly by enhancing initial local spreading. As a result, there is an optimal modularity at which both local and global spreading can occur~\cite{Nematzadeh2014a}.

In the context of neural dynamics, this result suggests that communities could offer a way to balance and arbitrate local and global communication and computation. We hypothesize that an ideal computing capacity emerges near the intersection between local cohesion and global connectivity, analogous to the optimal modularity for information diffusion. 

We test whether this can be true in reservoir computers. Reservoir computers are biologically plausible models for brain computation~\cite{Enel2016, Soriano2015, Yamazaki2007} as well as a successful machine learning paradigm~\cite{Lukosevicius2009}. They have emerged as an alternative to traditional recurrent neural network (RNN) paradigm~\cite{Jaeger2004, Maass2002a}.

Instead of training all the connection parameters as in RNNs, reservoir computers only train a small number of readout parameters.
Reservoir computers use the implicit computational capacities of a neural reservoir---a network of model neurons. 
Compared with other frameworks that require training numerous parameters, this paradigm allows for larger networks and better parameter scaling. Reservoir computers have been successful in a range of tasks including time series prediction, natural language processing, and pattern generation, and have also been used as biologically plausible models for neural computation~\cite{Yamazaki2007, Holzmann2010, Jaeger2012, Triefenbach2010,Jalalvand2016, Deng2016, Souahlia2016, Soriano2015, Rossert2015, Enel2016}.

Reservoir computers operate by taking an input signal(s) into a high-dimensional reservoir state space where signals are mixed. We use echo state networks (ESN)--a popular implementation of reservoir computing--where the reservoir is a collection of randomly connected neurons and the inputs are continuous or binary signals that are injected into a random subset of those neurons through randomly weighted connections. The reservoir's output is read via a layer of read-out neurons which receive connections from all neurons in the reservoir. They have no input back into the reservoir and they act as the system's output on tasks.

\begin{figure}[ht]
\includegraphics[width=\linewidth]{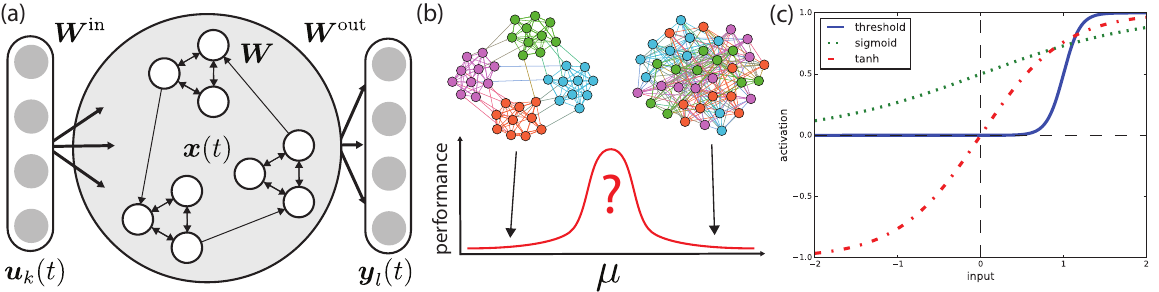}
\caption{(a) A modular echo state network (ESN). At each time step a $k$-dimensional input signal $\boldsymbol{u}_k(t)$ is introduced with randomly weighted input weights $\boldsymbol{W}^{\text{in}}$. The reservoir's state $x(t)$ evolves through a randomly-generated constant weight matrix $\boldsymbol{W}$.
The output weights $\boldsymbol{W}^{\text{out}}$ are trained based on the tasks. (b) $\mu$ is the fraction of bridges that connect communities within the reservoir. At low $\mu$ community structure is pronounced, while communities vanish at high $\mu$ ($\approx 0.5$). We hypothesize that performance increases when a balance between the local cohesion of communities and the global connectivity of bridges is met. (c) A visual comparison of activation functions. Our activation function (solid blue) has threshold-like behavior where small inputs invoke no response up to a threshold after which the neuron becomes excited. This type of activity mimics the kind expressed in many biological neural networks.}
\label{fig:esn}
\end{figure}

The reservoir weights and input weights are generally drawn from a given probability distribution and remain unchanged, while the read-out weights that connect the reservoir and read-outs are trained (see Fig.~\ref{fig:esn}(a)). Read-out neurons can be considered as ``tuning knobs'' into the desired set of non-linear computations that are being performed within the reservoir. Therefore, the ability of a reservoir computer to learn a particular behavior depends on the richness of the dynamical repertoire of the reservoir~\cite{Lukosevicius2009, Pascanu2011a}.

Many attempts have been made to calibrate reservoirs for particular tasks. In echo state networks this usually entails the adjustment of the spectral radius (largest eigenvalue of the reservoir weight matrix), the input and reservoir weight scales, and reservoir size~\cite{Jaeger2002, Pascanu2011a, Rodan2011, Farkas2016}.
In memory tasks, performance peaks sharply around a critical point for the spectral radius, whereby the neural network resides within a dynamical regime with long transients and ``echos'' of previous inputs reverberating through the states of the neurons preserving past information~\cite{Verstraeten2007, Pascanu2011a}. Weight distribution has also been found to play an important role in performance~\cite{Ju2013}, and the effects of reservoir topology has been studied using small-world~\cite{Deng2007}, scale-free~\cite{Deng2007}, columnar~\cite{Maass2002a, Verstraeten2007, Ju2013, Li2015}, Kronecker graphs~\cite{Rad2008, Leskovec2010}, and ensembles with lateral inhibition~\cite{Xue2007}, each of which showing improvements in performance over simple random graphs.

Echo state networks provide a compelling substrate for investigating the relationship between community structure, information diffusion, and memory. They can be biologically realistic, are simple to train; the separation between the reservoir and the trained readouts means that the training process does not interfere in the structure of the reservoir itself (see the Supplemental Material Table~\ref{table:tech}).

Here, we take a principled approach based on the theory of network structure and information diffusion to test a hypothesis that the best memory performance emerges when a neural reservoir is at the optimal modularity for information diffusion, where local and global communication can be easily balanced (see the Supplemental Material Fig.~\ref{fig:concept_diagram}).
We implement neural reservoirs with different levels of community structure (see Fig.~\ref{fig:esn}(a)) by fixing the total number of links and communities while adjusting a mixing parameter $\mu$ which controls the fraction of links between communities. Control of this parameter lets us explore how community structure plays a role in performance on two memory tasks (see Fig.~\ref{fig:esn}(b)).
Three simulations are performed. The first tests for the presence of the optimal modularity phenomena in the ESNs. The second uses the same ESNs to perform a memory capacity task to determine the relationship between the optimal modularity phenomena and task performance. Lastly, we investigate the relationship between community structure and the capacity of the ESN to recall unique patterns in a memorization task.

For the tasks we use a threshold-like activation function (see Fig.~\ref{fig:esn}(c)), which is a more biologically plausible alternative to the tanh or linear neurons often used in artificial neural networks.
The key distinction between the threshold-like activation function and tanh activation functions is that threshold-like functions only excite postsynaptic neurons if enough presynaptic neurons activate in unison. On the other-hand, postsynaptic tanh neurons will always activate in proportion to presynaptic neurons, no matter how weak those activations are.

\section{Results}

\subsection{Optimal modularity in reservoir dynamics}

\begin{figure}[ht]
\includegraphics[width=\linewidth]{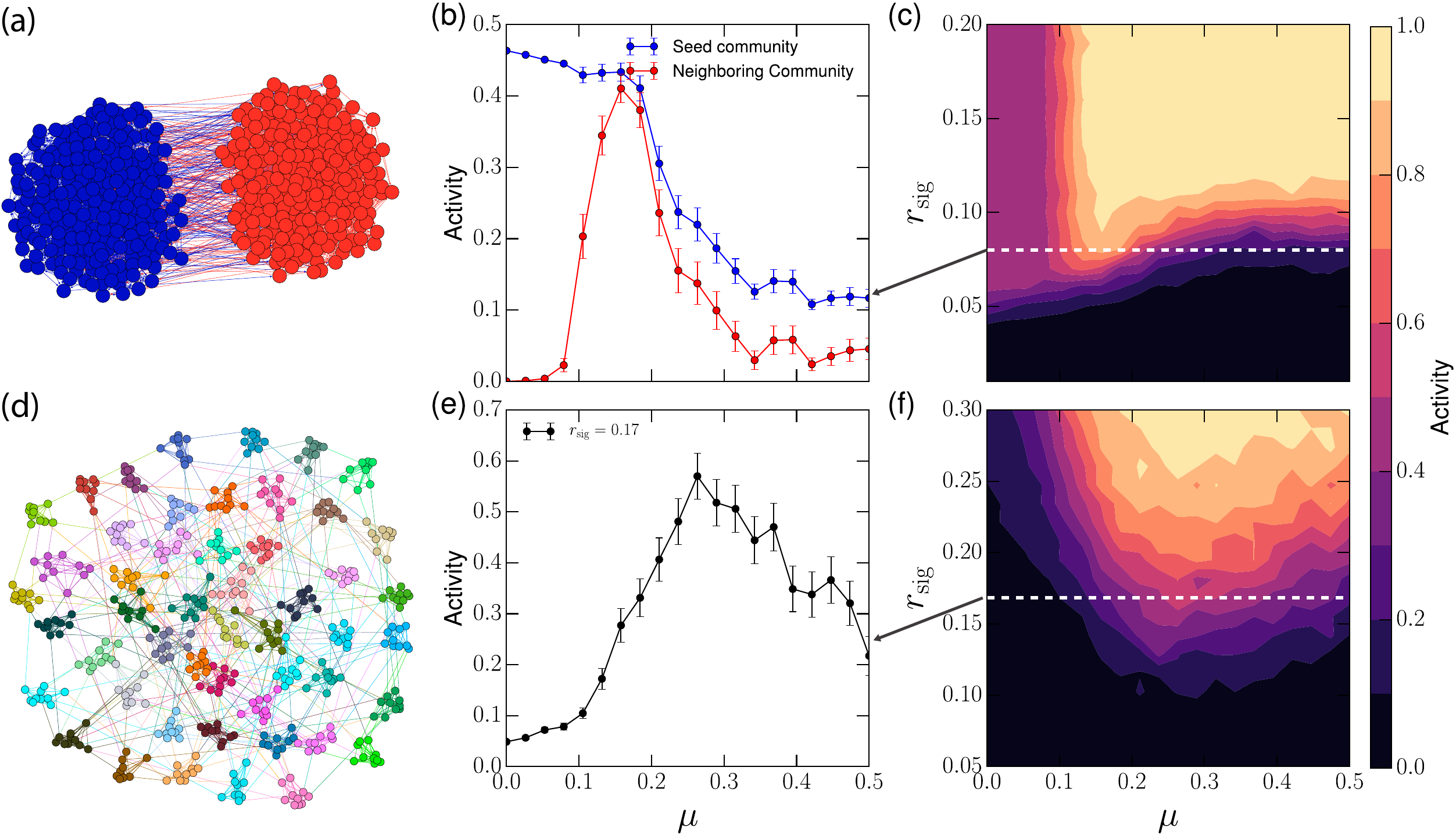}
\caption{(a) A two community network of threshold-like neurons receives input into the seed community (blue). (b) An optimal region with maximum activation emerges. (c) Phase diagram for the two-community case. Communities behave similar to gating functions, that can be turned on and transmit information once the input surpass a threshold. (d) Reservoirs with many communities and randomly injected input also exhibit optimal modularity. (e) The activity level of the network is shown. At low $\mu$ no single community receives enough signal to be activated, while at high $\mu$ internal cohesion is too weak to recruit other nodes. In between, the signal can be consolidated effectively, activating larger portions of the network. (f) The full phase-diagram showing the total fractional activity of the network. Error bars represent the standard error of the mean.}
\label{fig:optmod}
\end{figure}

We first test whether the optimal modularity phenomenon found in the linear threshold model can be generalized to neural reservoirs by running two simulations. Nodes governed by the linear threshold model remain active once turned on, and are not good units for computing. Instead we use a step-like activation function (see Fig.~\ref{fig:esn}(c)). First, we assume a simple two-community configuration as in the original study~\cite{Nematzadeh2014a}(see Fig.~\ref{fig:optmod}(a)), where the fraction of bridges $\mu$ controls the strength of community structure in the network. When $\mu=0$, the communities are maximally strong and disconnected, and when $\mu\approx0.5$ the community structure vanishes. The average degree and the total number of edges remains constant as $\mu$ is varied. An input signal is injected into a random fraction of the neurons ($r_{\text{sig}}$) in a seed community and the activity response of each community is measured. The results confirm the generalizability of the optimal modularity phenomenon for neural networks. 

At low $\mu$, strong local cohesion activates the seed community, while the neighboring community remains inactive as there are too few bridges (see Fig.~\ref{fig:optmod}(b)). At high $\mu$ there are enough bridges to transmit information globally but not enough internal connections to foster local spreading, resulting in a weak response. An optimal region emerges where local cohesion and global connectivity are balanced, maximizing the response of the whole network, as was demonstrated in \cite{Nematzadeh2014a} for linear threshold models. 
The fraction of neurons that receive input ($r_{\text{sig}}$) modulates the behavior of the communities. The phase diagram in Fig.~\ref{fig:optmod}(c) shows how the system can switch from being inactive at low $r_{\text{sig}}$, to a single active community, to full network activation as the fraction of activated neurons increases. The sharpness of this transition means the community behaves like a threshold-like function as well.
Though we control $r_{\text{sig}}$ as a static parameter in this model, it can represent the fraction of active neural pathways between communities, which may vary over time. Communities could switch between these inactive and active states in response to stimuli based on their activation threshold, allowing them to behave as information gates.

Our second study uses a more general setting, a reservoir with many communities similar to ones that might be used in an ESN or observed in the brain (see Fig.~\ref{fig:optmod}(d)). The previous study only examined input into a single community, here we extend that to many communities.
In Fig.~\ref{fig:optmod}(e) we record the response of a $50$-community network that receives a signal that is randomly distributed across the whole network. 
The result shows that even when there is no designated seed community, similar optimal modularity behavior arises.
At low $\mu$ the input signal cannot be reinforced due to the lack of bridges, and is unable to excite even the highly cohesive communities. 
At high $\mu$ the many global bridges help to consolidate the signal, but there is not enough local cohesion to continue to facilitate a strong response. 
In the optimal region there is a balance between the amplifying effect of the communities and the global communication of the bridges which enables the network to take a sub-threshold, globally distributed signal and spread it throughout the network. In linear and tanh reservoirs, no such relationship is found (see the Supplemental Material Fig.~\ref{fig:linear_activation_results} and Fig.~\ref{fig:tanh_activation_results}); instead communities behave in a more intuitive fashion, restricting information flow.

\subsection{Optimal modularity in a memory capacity task}
\begin{figure}
\centerline{\includegraphics[width=\textwidth]{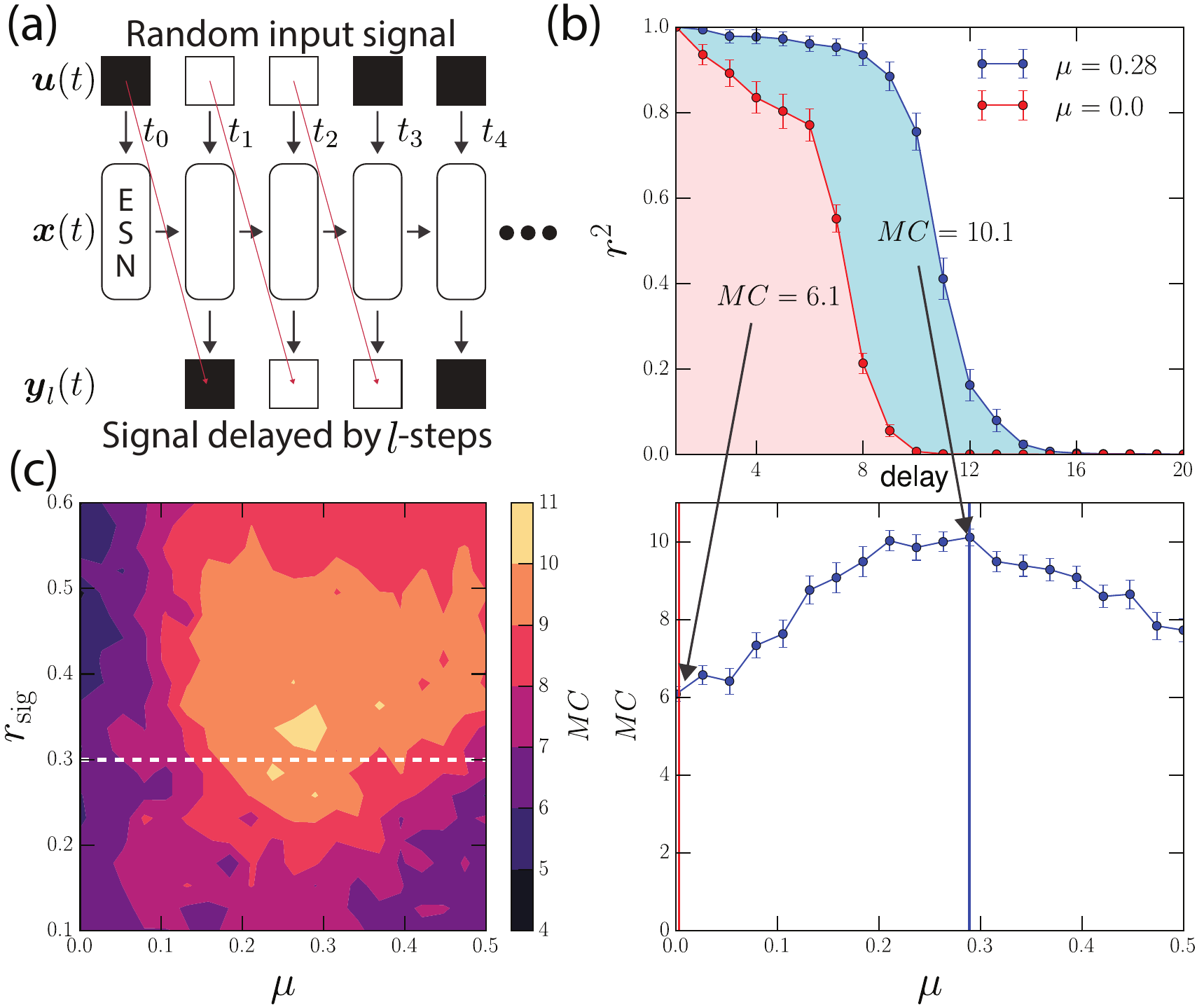}}
\caption{(a) A memory capacity task for measuring the memory duration of ESNs. Readout nodes are trained to reproduce a delayed input sequence. The delay varies from $1$ to $l$, where $l$ is the number of readouts. (b) Top: The performance is defined by the coefficient of determination ($r^2$) between the input signal and the output of the node. If the $r^2$ is $1.0$, then the readout perfectly reproduces the inputs. ${MC}$ denotes the overall performance of the ESN on the task. It represents the area under the curve of the $r^2$ versus delay plot (see shaded regions). (b) Bottom: The average performance over many reservoirs is shown as a function of $\mu$ where performance is maximal at intermediate levels of modularity. It is taken as a slice through (c) the complete contour-diagram for the task. Error bars represent the standard error of the mean.}
\label{fig:mctask}
\end{figure}

We test whether optimal modularity provides a benefit to the ESN's memory performance by a common memory benchmark task developed by Jaeger~\cite{Jaeger2002} (see Fig.~\ref{fig:mctask}(a)). The task involves feeding a stream of random inputs into the reservoir and training readout neurons to replay the stream at various time-lags. The coefficient of determination between the binomially distributed input signal and a delayed output signal for each delay parameter is used to quantify the performance of the ESN. The memory capacity ($\text{MC}$) of the network is the sum of these performances over all time-lags as shown by the shaded region in Fig.~\ref{fig:mctask}(b).

Reservoirs with strong community structure (low $\mu$) exhibit the poorest performance; the reservoirs are ensembles of effectively disconnected reservoirs, with little to no inter-community communication. Performance improves substantially with $\mu$ as the fraction of global bridges grows, facilitating inter-community communication. 
A turnover point is reached beyond which replacing connections with bridges compromises local cohesion. 
After a certain point, larger $\mu$ leads to performance loss. The region of elevated performance corresponds to the same region of optimal modularity on a reservoir with the same properties and inputs as those used in the task (see the Supplemental Material Fig.~\ref{fig:supp_overlap}).

We also examine the impact of input signal strength. In Fig.~\ref{fig:mctask}(c) we show that this optimal region of performance holds over a wide range of $r_{\text{sig}}$, and that there is a narrow band near $r_{\text{sig}}\approx 0.3$ where the highest performance is achieved around $\mu\approx 0.2$. As expected, we also see a region of optimal $r_{\text{sig}}$ for reservoirs, because either under- or over-stimulation is disadvantageous.
Yet, the added benefit of community structure is due to more than just the amplification of the signal.
If communities were only amplifying the input signal, then increasing $r_{\text{sig}}$ in random graphs should give the same performance as that found in the optimal region, but this is not the case. Fig.~\ref{fig:mctask}(c) shows that random graphs are unable to meet the performance gains provided near optimal $\mu$ regardless of $r_{\text{sig}}$. Additionally, this optimal region remains even if we control for changes in the spectral radius of the reservoir's adjacency matrix, which is known to play an important role in ESN memory capacity for linear and tanh systems~\cite{Jaeger2002, Verstraeten2007, Farkas2016} (see the Supplemental Material Fig.~\ref{fig:supp_ws}-\ref{fig:supp_spectral_rsig}). In such systems modularity reduces memory capacity, as communities create an information bottleneck (see the Supplemental Material Fig.~\ref{fig:linear_mc_results}-\ref{fig:tanh_mc_results}).
However, weight scale still plays a larger role in determining the level of performance for ESNs in our simulations (see the Supplemental Material Fig.~\ref{fig:supp_ws}). There is also a performance difference between the increasingly nonlinear activation functions, with linear performing best, and tanh and sigmoid performing worse; illustrating a previously established trade-off between memory and nonlinearity~\cite{Dambre2012, Verstraeten2010, Verstraeten2007}.
Lastly, ESN performance has been attributed to reservoir sparsity in the past~\cite{Jaeger2004, Lukosevicius2012a}, however as node degree, average node strength, and total number of edges remain constant as $\mu$ changes such effects are controlled for.

\subsection{Optimal modularity in a recall task}

\begin{figure}
\centerline{\includegraphics[width=\textwidth]{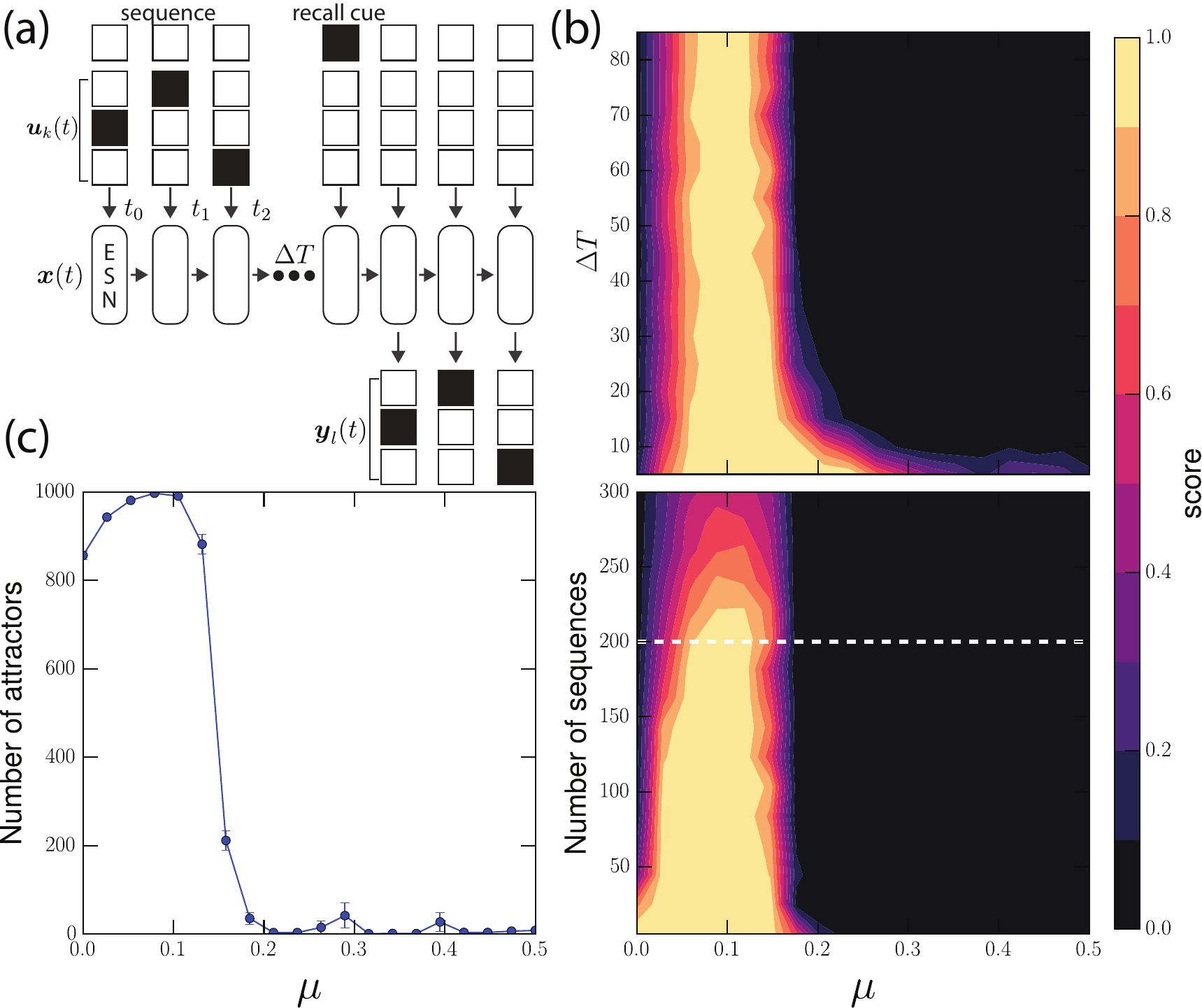}}
\caption{(a) A recall task for testing the amount of patterns that the ESN can learn. For this task, a randomly generated sequence of binary inputs across several dimensions are fed into the reservoir. After $\Delta T$ time steps, when it receives a cue, it must reproduce the original input sequence. The ESN is trained on each sequence. Performance on the recall task is determined by the fraction of perfect recalls from the learned sequences. A score of $1.0$ means that all learned sequences were correctly recalled. (b) Top: Performance is measured against $\Delta T$, displaying the maximal performance at $\mu\approx0.1$. (b) Bottom: The number of sequences that the ESNs can remember for long periods ($\Delta T=80$), shows a similar optimal region. (c) The best performing, optimally modular networks have many more available attractors. Error bars represent the standard error of the mean.}
\label{fig:recalltask}
\end{figure}

We employ another common memory task that estimates a different feature of memory: the number of unique patterns that can be learned. 
This requires a rich attractor space that can express and maintain many unique sequences. From here out we consider an attractor to be a basin of state (and input) configurations that lead to the same fixed point in the reservoirs state space.
In this task, a sequence of randomly generated $0$'s and $1$'s is fed to the network as shown in Fig.~\ref{fig:recalltask}(a).
For the simulation, we use sets of $4\times5$ dimensional binary sequences as input.
The readouts should then learn to recall the original sequence after an arbitrarily long delay $\Delta T$ and the presentation of a recall cue of 1 (for one time-step) through a separate input channel.

By varying $\mu$ we can show how recall performance changes with community structure. 
Fig.~\ref{fig:recalltask}(b) top shows the average performance measured by the fraction of perfectly recalled sequences, for a set of $200$ sequences. Well performing reservoirs are able to store the sequences in attractors for arbitrarily long times. Similar to the memory capacity task, we see the poorest performance for random networks and networks with low $\mu$. There is a sharp spike in performance near $\mu\approx0.1$.
The average performance over the number of sequences (when $\Delta T=80$) show that optimal performance at $\mu$ starts to drop off after $\approx230$ sequences (Fig.~\ref{fig:recalltask}(b) bottom).

We investigate the discrepancy in performance between modular and non-modular networks by examining the reservoir attractor space. 
We measure the number of unique available attractors that the reservoirs would be exposed to by initializing the reservoirs at initial conditions associated with the sequences we use. We find a skewed response from the network as shown in Fig.~\ref{fig:recalltask}(c) where the number of available attractors is maximized when $\mu > 0$. Many of these additional attractors between $0.0<\mu<0.2$ are limit cycles that result from the interaction between the communities in the reservoir.

The attractor space provides insights about the optimal region. At higher $\mu$ the whole reservoir behaves as a single system, leaving very few attractors for the network to utilize for information storage. The reservoir has to rely on short-lived transients for storage. With extremely modular structure ($\mu\approx0$), reservoirs have the most available attractors, but they are not readily discriminated by the linear readouts. Surprisingly, these attractors are more readily teased apart as communities become more interconnected. However, there is a clear trade-off, as too much interconnection folds all the initial conditions into a few large attractor basins.

\section{Discussion}

Biological neural networks are often modeled using neurons with threshold-like behavior, such as integrate-and-fire neurons, the Grossberg-Cohen model, or Hopfield networks. Reservoirs of threshold-like neurons, like those presented here, provide a simple model for investigating the computational capabilities of biological neural networks. By adopting and systematically varying topological characteristics akin to those found in brain networks, such as modularity, and subjecting those networks to tasks we can gain insight into the functional advantages provided by these architectures.

We have demonstrated that ESNs exhibit optimal modularity both in the context of signal spreading and memory capacity and they are closely linked to the optimal modularity for information spreading. 
Through dynamical analysis we found that balancing local and global cohesion enabled modular reservoirs to spread information across the network and consolidate distributed signals, although alternative mechanisms may also be in play, such as cycle properties~\cite{Garcia2014}.
We then showed that such optimal regions coincide with the optimal community strength that exhibit the best memory performance.
Both the memory capacity and recall task benefited by adopting modular structures over random networks, despite performing in different dynamical regimes (equilibrium versus non-equilibrium).

A key component of our hypothesis is the adoption of a threshold-like (or step-like) activation function for our ESNs, which is a more biologically plausible alternative to the tanh or linear neurons often used in artificial neural networks. The optimal modularity phenomenon emerges only for neural networks of threshold-like neurons and does not exist for neural networks of linear or tanh neurons (i.e. simple contagions) used in traditional ESNs, and so many developed intuitions about ESN dynamics and performance may not readily map to ESNs driven by complex contagions like the ones here. Indeed, the relationship between network topology and performance is known to vary with the activation function, with threshold-like or spiking neurons (common in liquid state machines~\cite{Maass2002a}) being more heavily dependent on topology~\cite{Bertschinger2004, Haeusler2007, Schrauwen2009}. Because the effects of modularity vary depending upon the activation function, a suitable information diffusion analysis should be chosen to explore the impact of network topology for a given type of spreading process. Moreover, because the benefits of modularity are specific to threshold-like neurons, distinct network design principles are needed for biological neural networks and the artificial neural networks used in machine learning. Additionally, as we have seen that the choice of architecture can have a profound impact on the dynamical properties that can emerge from the neural network, there maybe value in applying these insights to the architectural design of recurrent neural networks in machine learning, where all weights in the network undergo training but where architecture is usually fixed.

While weight scale remains the most important feature of the system in determining performance, our results suggest significant computational benefits of community structure, and contributes to understanding the role it plays in biological neural networks~\cite{Hilgetag2000, Bullmore2009, Meunier2010, Sporns2004,Hagmann2008, Shimono2015a, Buxhoeveden2002, Constantinidis2016} which are also driven by complex contagions and possess modular topologies. The dynamical principles of information spreading mark trade-offs in the permeability of information on the network that can promote or hinder performance. While this analysis provides us some insight, it remains an open question as to whether our results can be generalized to the context of more realistic biological neural networks where spike-timing dependent plasticity and neuromodulation play a key role in determining the network's dynamical and topological characteristics.

In addition to the optimal region and the ability of communities to foster information spreading and improved performance among threshold-like neurons, modularity may play other important roles. For instance, it offers a way to compartmentalize advances and make them robust to noise (e.g. the watchmaker's parable~\cite{Simon1997}). Modularity also appears to confer advantages to neural networks in changing environments~\cite{Kashtan2005}, under wiring cost constraints~\cite{Clune2013}, when learning new skills~\cite{Ellefsen2015}, and under random failures~\cite{Kaiser2004}.
These suggest additional avenues for exploring the computational benefits of modular reservoirs and neural networks. And it is still an open question how community structure affects performance on other tasks like signal processing, prediction, or system modeling.

Neural reservoirs have generally been considered ``black-boxes'', yet through combining dynamical, informational, and computational studies it maybe possible to build a taxonomy of the functional implications of topological features for both artificial and biological neural networks. Dynamical and performative analysis of neural networks can afford valuable insights into their computational capabilities as we have seen here.

\section{Methods}
Our ESN architecture with community structure is shown in Fig.~\ref{fig:esn}(a). 
The inputs are denoted as $\boldsymbol{u}_k(t)$ which is a $k$-dimensional vector. 
Each dimension of input is connected to a random subset of neurons in the reservoir.
$\boldsymbol{x}(t)$ is the $N$-dimensional state vector of the reservoir, where $N$ is the number of reservoir neurons. 
$\boldsymbol{y}_l(t)$ represents the states of the $l$ readout neurons. 
The $k$ inputs are connected by a $N\times k$ matrix $\boldsymbol{W}^{\textup{in}}$ to the $N$ neurons.
The network structure of the reservoir is represents by a $N\times N$ weight matrix $\boldsymbol{W}$, and the output weights are represented by an $N\times l$ matrix $\boldsymbol{W}^{\textup{out}}$.
The reservoirs follow the standard ESN dynamics without feedback or time constants:

\begin{eqnarray}
\boldsymbol{x}(t+1) = f\left( \boldsymbol{W} \boldsymbol{x}(t) + \boldsymbol{W}^{\textup{in}} \boldsymbol{u}(t+1) \right)\text{,}
\end{eqnarray}

\begin{eqnarray}
\boldsymbol{y}(t)=g\left( \boldsymbol{W}^{\textup{out}} \left[ \boldsymbol{x}(t) :\boldsymbol{u}(t) \right] \right)\text{.}
\end{eqnarray}

Here $f$ is the reservoir activation function, $g$ is the readout activation function, and $[a:b]$ denotes the concatenation of two vectors. Often $f$ is chosen to be a sigmoid-like function such as tanh, while $g$ is often taken to be linear~\cite{Lukosevicius2009}. However in our case we use a general sigmoid function:

\begin{eqnarray}
f(z)= \frac{a}{b+e^{-k(z-c)}}-d\text{,}
\end{eqnarray}

with parameters $a=1$, $b=1$, $c=1$, $k=10$, and $d=0$ giving a non-linear threshold-like activation function making it step-like in shape and a \emph{complex contagion} like other neuron models (e.g. integrate-and-fire, Hopfield, or Wilson-Cowan models). For the readout neurons, $g$ is chosen to be a step function:

\begin{eqnarray}
g(z)=\begin{cases}
0 & z\leq 0.5 \text{,}\\
1 & z> 0.5 \text{.}
\end{cases}
\end{eqnarray}

Linear regression is used to solve for $\boldsymbol{W}^{\text{out}}$. $\boldsymbol{W}^{\textup{out}}=\boldsymbol{Y}^{\textup{tar}}\boldsymbol{X}^{+}$ where $\boldsymbol{Y}^{\textup{tar}}$ is an $l \times T$ matrix of target outputs over a time course $T$, and $\boldsymbol{X}^{+}$ is the pseudo-inverse of the history of the reservoir state vector (where $\boldsymbol{X} \in \mathbb{R}^{N \times T} $) ~\cite{Lukosevicius2009}.
To generate the reservoirs we use the LFR benchmark model~\cite{Lancichinetti2008}, which can generate random graphs with a variety of community structures. The LFR benchmark model uses a configuration model to generate random graphs. The configuration model works by imposing a degree sequence to the nodes and randomly wiring the edge ``stubs''~\cite{Newman2010B}. The LFR model extends this by including community assignment and rewiring steps to constrain the fraction of bridges in the network. Due to its relationship with the configuration model, LFR graphs exhibit low average shortest path length and low average clustering coefficient in contrast to the Wattz-Strogatz models that have low average shortest path length and high clustering. For small graphs like the ones we use for building reservoirs, the average shortest path length increases monotonically with decreasing $\mu$. This is due to the sparseness of directed links between communities. As $\mu$ approaches 0 the communities become disconnected. In our case we vary the fraction of bridges ($\mu$) in the network while holding the degree distribution and total number of edges the same, controlling for the density of connections in the network. Weights for the network are drawn separately from a uniform distribution and described in following sections. Code for all the simulations and tasks is available online.

\subsection{Reservoir dynamics}

We used reservoirs with $N=500$ nodes with every node having a degree of $6$. Reservoir states were initialized with a zero vector, $\boldsymbol{x}(0)=\lbrace 0,\dots,0\rbrace$. The first experiment uses a two community cluster of $250$ nodes each, matching the scenario from \cite{Nematzadeh2014a}.
Input was injected into $r_{\text{sig}}$ fraction of neurons into the seed community. 
The input signal lasted for the duration of the task until the system reached equilibrium at time $t_e$.
The final activation values of the neurons were summed within each community and used to calculate the fractional activation of the network for each community shown in Fig.~\ref{fig:optmod}(b); where the mean over $48$ reservoir realizations is shown. All activations were summed and divided by the size of the network to give the total fractional activation $1/N \sum^{N}_{i=1}x_i(t_e)$ as shown in Fig.~\ref{fig:optmod}(c).

In the following experiment, a reservoir of the same size, but with $50$ communities with $10$ nodes each was used. This time however, the input signal was not limited to a single community but applied randomly to nodes across the network. Again the signal was active for the full duration of the task until the system reached equilibrium when the final activation values of the neurons were summed within each community. Fig.~\ref{fig:optmod}(e) shows the activation for each community averaged over $48$ reservoir realizations and the total fractional activity in the network is then shown in Fig.~\ref{fig:optmod}(f).

Different measures for information spreading produce similar results. Also, optimal spreading can be observed in the transitory dynamics of the system, such as in networks that receive short input bursts and return to an inactive equilibrium state. Optimality for step-like activations has been shown to emerge regardless of community or network size using message-passing approximations~\cite{azadeh2018}. For many-community cases with distributed input, optimality existence in infinite networks depends upon community variation (e.g. size, edge density, number of inputs).

\subsection{Memory Capacity task}

The memory capacity task involves the input of a random sequence of numbers that the readout neurons are then trained on at various lags (see Fig.~\ref{fig:mctask}). There is just one input dimension and values of zero and one are input into a fraction of the reservoir's neurons $r_{\text{sig}}$. For each time lag there is a set of readout neurons that are trained independently to remember the input at the given time lag. The readout-neurons that maximize the coefficient of determination (or the square of the correlation coefficient) between the input signal and lagged output are used as the $k$-th delayed short-term memory capacity of the network $\mathrm{MC}_k$. The MC of the ESN becomes the sum over all delays:

\begin{eqnarray}
\mathrm{MC} = \sum_{k=1}^{\infty} \mathrm{MC}_k = \sum_{k=1}^{\infty}\frac{\text{cov}(u(t-k),y_k(t))^2}{\text{var}(u(t))\text{var}(y_k(t))}
\end{eqnarray}

We operationalize this sum as the memory capacity of the network. Unlike Jaeger's task, we input a binomial distribution of 1's and 0's rather than continuous values (see Fig.~\ref{fig:mctask}(a)). We try to keep the network small enough and sparse enough to reduce computational load, while still being large enough to solve the task. A reservoir of $N=500$ nodes and $50$ communities of size $10$ were used. Every node has a degree of $6$. The degree was chosen to be sparse enough to help reduce computing time, while high enough to support a wide range of modularities, which are partly constrained by degree. Reservoir parameters were not fitted to the task, rather a grid search was executed to find parameter sets that performed well, as the focus of the experiment is not to break records on memory performance, but rather to see how it changes with modularity. Among the parameters adjusted were the upper and lower bounds of the weight distribution and the weight scale ($W_s$) which adjusts the strengths of all the reservoir weights by a scalar value. Performance over the full range of $\mu$ values was evaluated at each point on the grid. Well performing reservoirs were found with weights between $-0.2$ and $1$ and with a weight scale parameter of $W_s=1.13$. The same was done for the input weight matrix where $\boldsymbol{W}^{\text{in}}$ also varies from $-0.2$ to $1$ with an input gain of $W_I=1.0$. Many viable parameters existed throughout the space which exhibit optimality. This is due in part to parameter coupling, where changing multiple parameters results in the same dynamics.

Each reservoir's read-outs were trained over a $1,500$ step sequence following the first $500$ steps that are removed to allow initial transients to die out. Once trained a new validation sequence of the same length is used to evaluate the performance of the ESN. Results averaged over $64$ reservoir samples are shown in Fig.~\ref{fig:mctask}(b,c). We also show the contour over $r_{\text{sig}}$ which is an important parameter in determining the performance of the reservoir. Performance peaks between $r_{\text{sig}}=0.3$ and $r_{\text{sig}}=0.4$ at a $\mu\approx0.25$.

\subsection{Recall task}

The recall task is a simplified version of the memory task developed by Jaeger~\cite{Jaeger2012}. A pattern of zeros and ones is input into the network which must recall that pattern after a distractor period. The ESN is trained on the whole set of unique sequences and the performance of the ESN is determined from its final output during the recall period, which occurs after the distractor period. We do this to estimate the total number of sequences that an ESN can remember. So unlike the memory capacity task that estimates memory duration given an arbitrary input sequence, the recall task quantifies the number of distinct signals an ESN can differentiate. 
This involves training an ESN on a set of sequences and then having it recall the sequences perfectly after a time delay $\Delta T$.
The input is a random $4\times 5$ binary set of 0's and 1's. At a single time step just one of the four input dimensions are active. This is in order to maintain the same level of external excitation per time-step, as we are not testing the network's dynamic range. The reservoir is initialized to a zero vector and provided with a random sequence. Following the delay period, a binary cue with value 1.0 is presented via a 5th input dimension. After this cue, the reservoir's read-out neurons must reproduce the input sequence. The read-out weights are trained on this sequence set. Fig.~\ref{fig:recalltask}(b) shows the average performance over $48$ reservoir samples. Many networks around the optimal $\mu$ value can retain the information for arbitrarily long times, as the task involves storing the information in a unique attractor. Fig.~\ref{fig:recalltask}(b) shows the average performance when $\Delta T=80$ as we vary the number of sequences. In Fig.~\ref{fig:recalltask}(c) we determine the average number of available attractors given inputs drawn from the full set of $4\times 5$ binary sequences where only one dimension of the input is active at a given time. For each of the $4\times 5$ binary sequences, the system was run until it reached the cue time, where a decision would be made by the readout layer. At this point converged trajectories would result in a failure to differentiate patterns. Two converged trajectories are determined to fall into the same attractor if the Euclidean distance between the system’s states are smaller than a value $\epsilon=0.1$. The number of attractor states is the number of these unique groupings and was robust to changes in $\epsilon$.
Parameters for the reservoir are chosen via a grid-search, as before, to find reasonable performance from which to start our analysis. Here reservoirs of size $N=1,000$ with node degree $7$ and community size $10$ are used. A larger reservoir was necessary in order to attain high performance on the task. Similarly, the weight distribution parameters are included in the search and reasonable performing reservoirs were found with weights drawn between $-0.1$ and $1.0$ with $W_s=1.0$, $r_{\text{sig}}=0.3$, an input gain of $W_I=2.0$ and uniform input weights of $1.0$. 

\section{Supportive Information}
Code can be found at \href{https://github.com/Nathaniel-Rodriguez/reservoirlib}{https://github.com/Nathaniel-Rodriguez/reservoirlib}.

\section{Acknowledgements}
We would like to thank John Beggs, Alessandro Flamini, Azadeh Nematzadeh, Pau Vilimelis Aceituno, Naoki Masuda, and Mikail Rubinov for helpful discussions and valuable feedback. This research was supported in part by Lilly Endowment, Inc., through its support for the Indiana University Pervasive Technology Institute, and in part by the Indiana METACyt Initiative. The Indiana METACyt Initiative at IU was also supported in part by Lilly Endowment, Inc. The Indiana University HPC infrastructure (Big Red II) helped make this research possible.

\section{Author Contributions}
All authors contributed to project formulation, theory, analysis, and writing. Additionally, NR contributed to coding, plotting, and simulation.

\bibliography{library}

\begin{thebibliography}{73}%
\makeatletter
\providecommand \@ifxundefined [1]{%
 \@ifx{#1\undefined}
}%
\providecommand \@ifnum [1]{%
 \ifnum #1\expandafter \@firstoftwo
 \else \expandafter \@secondoftwo
 \fi
}%
\providecommand \@ifx [1]{%
 \ifx #1\expandafter \@firstoftwo
 \else \expandafter \@secondoftwo
 \fi
}%
\providecommand \natexlab [1]{#1}%
\providecommand \enquote  [1]{``#1''}%
\providecommand \bibnamefont  [1]{#1}%
\providecommand \bibfnamefont [1]{#1}%
\providecommand \citenamefont [1]{#1}%
\providecommand \href@noop [0]{\@secondoftwo}%
\providecommand \href [0]{\begingroup \@sanitize@url \@href}%
\providecommand \@href[1]{\@@startlink{#1}\@@href}%
\providecommand \@@href[1]{\endgroup#1\@@endlink}%
\providecommand \@sanitize@url [0]{\catcode `\\12\catcode `\$12\catcode
  `\&12\catcode `\#12\catcode `\^12\catcode `\_12\catcode `\%12\relax}%
\providecommand \@@startlink[1]{}%
\providecommand \@@endlink[0]{}%
\providecommand \url  [0]{\begingroup\@sanitize@url \@url }%
\providecommand \@url [1]{\endgroup\@href {#1}{\urlprefix }}%
\providecommand \urlprefix  [0]{URL }%
\providecommand \Eprint [0]{\href }%
\providecommand \doibase [0]{http://dx.doi.org/}%
\providecommand \selectlanguage [0]{\@gobble}%
\providecommand \bibinfo  [0]{\@secondoftwo}%
\providecommand \bibfield  [0]{\@secondoftwo}%
\providecommand \translation [1]{[#1]}%
\providecommand \BibitemOpen [0]{}%
\providecommand \bibitemStop [0]{}%
\providecommand \bibitemNoStop [0]{.\EOS\space}%
\providecommand \EOS [0]{\spacefactor3000\relax}%
\providecommand \BibitemShut  [1]{\csname bibitem#1\endcsname}%
\let\auto@bib@innerbib\@empty
%</preamble>
\bibitem [{\citenamefont {LeCun}\ \emph {et~al.}(2015)\citenamefont {LeCun},
  \citenamefont {Bengio},\ and\ \citenamefont {Hinton}}]{LeCun2015}%
  \BibitemOpen
  \bibfield  {author} {\bibinfo {author} {\bibfnamefont {Y.}~\bibnamefont
  {LeCun}}, \bibinfo {author} {\bibfnamefont {Y.}~\bibnamefont {Bengio}}, \
  and\ \bibinfo {author} {\bibfnamefont {G.}~\bibnamefont {Hinton}},\ }\href
  {\doibase 10.1038/nature14539} {\bibfield  {journal} {\bibinfo  {journal}
  {Nature}\ }\textbf {\bibinfo {volume} {521}},\ \bibinfo {pages} {436}
  (\bibinfo {year} {2015})}\BibitemShut {NoStop}%
\bibitem [{\citenamefont {Hubel}\ and\ \citenamefont
  {Wiesel}(1972)}]{Wiesel1972}%
  \BibitemOpen
  \bibfield  {author} {\bibinfo {author} {\bibfnamefont {D.~H.}\ \bibnamefont
  {Hubel}}\ and\ \bibinfo {author} {\bibfnamefont {T.~N.}\ \bibnamefont
  {Wiesel}},\ }\href {\doibase 10.1002/cne.901460402} {\bibfield  {journal}
  {\bibinfo  {journal} {J. Comp. Neurol.}\ }\textbf {\bibinfo {volume} {146}},\
  \bibinfo {pages} {421} (\bibinfo {year} {1972})}\BibitemShut {NoStop}%
\bibitem [{\citenamefont {Otmakhova}\ \emph {et~al.}(2013)\citenamefont
  {Otmakhova}, \citenamefont {Duzel}, \citenamefont {Deutch},\ and\
  \citenamefont {Lisman}}]{Baldassarre2013}%
  \BibitemOpen
  \bibfield  {author} {\bibinfo {author} {\bibfnamefont {N.}~\bibnamefont
  {Otmakhova}}, \bibinfo {author} {\bibfnamefont {E.}~\bibnamefont {Duzel}},
  \bibinfo {author} {\bibfnamefont {A.~Y.}\ \bibnamefont {Deutch}}, \ and\
  \bibinfo {author} {\bibfnamefont {J.}~\bibnamefont {Lisman}},\ }in\
  \href@noop {} {\emph {\bibinfo {booktitle} {Intrinsically Motiv. Learn. Nat.
  Artif. Syst.}}},\ Vol.\ \bibinfo {volume} {9783642323}\ (\bibinfo
  {publisher} {Springer Berlin Heidelberg},\ \bibinfo {address} {Berlin,
  Heidelberg},\ \bibinfo {year} {2013})\ pp.\ \bibinfo {pages}
  {235--254}\BibitemShut {NoStop}%
\bibitem [{\citenamefont {Ioffe}\ and\ \citenamefont
  {Szegedy}(2015)}]{Ioffe2015}%
  \BibitemOpen
  \bibfield  {author} {\bibinfo {author} {\bibfnamefont {S.}~\bibnamefont
  {Ioffe}}\ and\ \bibinfo {author} {\bibfnamefont {C.}~\bibnamefont
  {Szegedy}},\ }in\ \href {\doibase 10.1007/s13398-014-0173-7.2} {\emph
  {\bibinfo {booktitle} {Proc. 32nd Int. Conf. Mach. Learn.}}}\ (\bibinfo
  {publisher} {JMLR Workshop and Conference Proceedings},\ \bibinfo {year}
  {2015})\ pp.\ \bibinfo {pages} {448--456}\BibitemShut {NoStop}%
\bibitem [{\citenamefont {Schmidhuber}(2015)}]{Schmidhuber2015}%
  \BibitemOpen
  \bibfield  {author} {\bibinfo {author} {\bibfnamefont {J.}~\bibnamefont
  {Schmidhuber}},\ }\href {\doibase 10.1016/j.neunet.2014.09.003} {\bibfield
  {journal} {\bibinfo  {journal} {Neural Networks}\ }\textbf {\bibinfo {volume}
  {61}},\ \bibinfo {pages} {85} (\bibinfo {year} {2015})}\BibitemShut {NoStop}%
\bibitem [{\citenamefont {Legenstein}\ and\ \citenamefont
  {Maass}(2005)}]{Legenstein2005}%
  \BibitemOpen
  \bibfield  {author} {\bibinfo {author} {\bibfnamefont {R.}~\bibnamefont
  {Legenstein}}\ and\ \bibinfo {author} {\bibfnamefont {W.}~\bibnamefont
  {Maass}},\ }\href@noop {} {\bibfield  {journal} {\bibinfo  {journal} {New
  Dir. Stat. Signal Process. From Syst. to Brain}\ ,\ \bibinfo {pages} {127}}
  (\bibinfo {year} {2005})}\BibitemShut {NoStop}%
\bibitem [{\citenamefont {Sussillo}\ and\ \citenamefont
  {Barak}(2013)}]{Sussillo2013}%
  \BibitemOpen
  \bibfield  {author} {\bibinfo {author} {\bibfnamefont {D.}~\bibnamefont
  {Sussillo}}\ and\ \bibinfo {author} {\bibfnamefont {O.}~\bibnamefont
  {Barak}},\ }\href@noop {} {\bibfield  {journal} {\bibinfo  {journal} {Neural
  Comput.}\ }\textbf {\bibinfo {volume} {25}},\ \bibinfo {pages} {626}
  (\bibinfo {year} {2013})}\BibitemShut {NoStop}%
\bibitem [{\citenamefont {Girvan}\ and\ \citenamefont
  {Newman}(2002)}]{Girvan2002}%
  \BibitemOpen
  \bibfield  {author} {\bibinfo {author} {\bibfnamefont {M.}~\bibnamefont
  {Girvan}}\ and\ \bibinfo {author} {\bibfnamefont {M.~E.~J.}\ \bibnamefont
  {Newman}},\ }\href {\doibase 10.1073/pnas.122653799} {\bibfield  {journal}
  {\bibinfo  {journal} {Proc. Natl. Acad. Sci.}\ }\textbf {\bibinfo {volume}
  {99}},\ \bibinfo {pages} {7821} (\bibinfo {year} {2002})}\BibitemShut
  {NoStop}%
\bibitem [{\citenamefont {Bullmore}\ and\ \citenamefont
  {Sporns}(2009)}]{Bullmore2009}%
  \BibitemOpen
  \bibfield  {author} {\bibinfo {author} {\bibfnamefont {E.~T.}\ \bibnamefont
  {Bullmore}}\ and\ \bibinfo {author} {\bibfnamefont {O.}~\bibnamefont
  {Sporns}},\ }\href {\doibase 10.1038/nrn2575} {\bibfield  {journal} {\bibinfo
   {journal} {Nat. Rev. Neurosci.}\ }\textbf {\bibinfo {volume} {10}},\
  \bibinfo {pages} {186} (\bibinfo {year} {2009})}\BibitemShut {NoStop}%
\bibitem [{\citenamefont {M{\"{u}}ller-Linow}\ \emph
  {et~al.}(2008)\citenamefont {M{\"{u}}ller-Linow}, \citenamefont {Hilgetag},\
  and\ \citenamefont {H{\"{u}}tt}}]{Muller-Linow2008}%
  \BibitemOpen
  \bibfield  {author} {\bibinfo {author} {\bibfnamefont {M.}~\bibnamefont
  {M{\"{u}}ller-Linow}}, \bibinfo {author} {\bibfnamefont {C.~C.}\ \bibnamefont
  {Hilgetag}}, \ and\ \bibinfo {author} {\bibfnamefont {M.-T.}\ \bibnamefont
  {H{\"{u}}tt}},\ }\href {\doibase 10.1371/journal.pcbi.1000190} {\bibfield
  {journal} {\bibinfo  {journal} {PLoS Comput. Biol.}\ }\textbf {\bibinfo
  {volume} {4}},\ \bibinfo {pages} {e1000190} (\bibinfo {year}
  {2008})}\BibitemShut {NoStop}%
\bibitem [{\citenamefont {Kaiser}\ and\ \citenamefont
  {Hilgetag}(2010)}]{Kaiser2010}%
  \BibitemOpen
  \bibfield  {author} {\bibinfo {author} {\bibfnamefont {M.}~\bibnamefont
  {Kaiser}}\ and\ \bibinfo {author} {\bibfnamefont {C.~C.}\ \bibnamefont
  {Hilgetag}},\ }\href {\doibase 10.3389/fninf.2010.00008} {\bibfield
  {journal} {\bibinfo  {journal} {Front Neuroinformatics}\ }\textbf {\bibinfo
  {volume} {4}},\ \bibinfo {pages} {8} (\bibinfo {year} {2010})}\BibitemShut
  {NoStop}%
\bibitem [{\citenamefont {Wang}\ \emph {et~al.}(2011)\citenamefont {Wang},
  \citenamefont {Hilgetag},\ and\ \citenamefont {Zhou}}]{Wang2011}%
  \BibitemOpen
  \bibfield  {author} {\bibinfo {author} {\bibfnamefont {S.-J.}\ \bibnamefont
  {Wang}}, \bibinfo {author} {\bibfnamefont {C.~C.}\ \bibnamefont {Hilgetag}},
  \ and\ \bibinfo {author} {\bibfnamefont {C.}~\bibnamefont {Zhou}},\ }\href
  {\doibase 10.3389/fncom.2011.00030} {\bibfield  {journal} {\bibinfo
  {journal} {Front. Comput. Neurosci.}\ }\textbf {\bibinfo {volume} {5}},\
  \bibinfo {pages} {1} (\bibinfo {year} {2011})}\BibitemShut {NoStop}%
\bibitem [{\citenamefont {Moretti}\ and\ \citenamefont
  {Mu{\~{n}}oz}(2013)}]{Moretti2013}%
  \BibitemOpen
  \bibfield  {author} {\bibinfo {author} {\bibfnamefont {P.}~\bibnamefont
  {Moretti}}\ and\ \bibinfo {author} {\bibfnamefont {M.~A.}\ \bibnamefont
  {Mu{\~{n}}oz}},\ }\href {\doibase 10.1038/ncomms3521} {\bibfield  {journal}
  {\bibinfo  {journal} {Nat. Commun.}\ }\textbf {\bibinfo {volume} {4}},\
  \bibinfo {pages} {2521} (\bibinfo {year} {2013})}\BibitemShut {NoStop}%
\bibitem [{\citenamefont {Villegas}\ \emph {et~al.}(2015)\citenamefont
  {Villegas}, \citenamefont {Moretti},\ and\ \citenamefont
  {Mu{\~{n}}oz}}]{Villegas2014}%
  \BibitemOpen
  \bibfield  {author} {\bibinfo {author} {\bibfnamefont {P.}~\bibnamefont
  {Villegas}}, \bibinfo {author} {\bibfnamefont {P.}~\bibnamefont {Moretti}}, \
  and\ \bibinfo {author} {\bibfnamefont {M.~A.}\ \bibnamefont {Mu{\~{n}}oz}},\
  }\href {\doibase 10.1038/srep05990} {\bibfield  {journal} {\bibinfo
  {journal} {Sci. Rep.}\ }\textbf {\bibinfo {volume} {4}},\ \bibinfo {pages}
  {5990} (\bibinfo {year} {2015})}\BibitemShut {NoStop}%
\bibitem [{\citenamefont {Boerlin}\ and\ \citenamefont
  {Den{\`{e}}ve}(2011)}]{Boerlin2011}%
  \BibitemOpen
  \bibfield  {author} {\bibinfo {author} {\bibfnamefont {M.}~\bibnamefont
  {Boerlin}}\ and\ \bibinfo {author} {\bibfnamefont {S.}~\bibnamefont
  {Den{\`{e}}ve}},\ }\href {\doibase 10.1371/journal.pcbi.1001080} {\bibfield
  {journal} {\bibinfo  {journal} {PLoS Comput. Biol.}\ }\textbf {\bibinfo
  {volume} {7}},\ \bibinfo {pages} {e1001080} (\bibinfo {year}
  {2011})}\BibitemShut {NoStop}%
\bibitem [{\citenamefont {Constantinidis}\ and\ \citenamefont
  {Klingberg}(2016)}]{Constantinidis2016}%
  \BibitemOpen
  \bibfield  {author} {\bibinfo {author} {\bibfnamefont {C.}~\bibnamefont
  {Constantinidis}}\ and\ \bibinfo {author} {\bibfnamefont {T.}~\bibnamefont
  {Klingberg}},\ }\href {\doibase 10.1038/nrn.2016.43} {\bibfield  {journal}
  {\bibinfo  {journal} {Nat. Rev. Neurosci.}\ }\textbf {\bibinfo {volume}
  {17}},\ \bibinfo {pages} {438} (\bibinfo {year} {2016})}\BibitemShut
  {NoStop}%
\bibitem [{\citenamefont {Cossart}\ \emph {et~al.}(2003)\citenamefont
  {Cossart}, \citenamefont {Aronov},\ and\ \citenamefont
  {Yuste}}]{Cossart2003}%
  \BibitemOpen
  \bibfield  {author} {\bibinfo {author} {\bibfnamefont {R.}~\bibnamefont
  {Cossart}}, \bibinfo {author} {\bibfnamefont {D.}~\bibnamefont {Aronov}}, \
  and\ \bibinfo {author} {\bibfnamefont {R.}~\bibnamefont {Yuste}},\ }\href
  {\doibase 10.1038/nature01614} {\bibfield  {journal} {\bibinfo  {journal}
  {Nature}\ }\textbf {\bibinfo {volume} {423}},\ \bibinfo {pages} {283}
  (\bibinfo {year} {2003})}\BibitemShut {NoStop}%
\bibitem [{\citenamefont {Klinshov}\ \emph {et~al.}(2014)\citenamefont
  {Klinshov}, \citenamefont {Teramae}, \citenamefont {Nekorkin},\ and\
  \citenamefont {Fukai}}]{Klinshov2014}%
  \BibitemOpen
  \bibfield  {author} {\bibinfo {author} {\bibfnamefont {V.~V.}\ \bibnamefont
  {Klinshov}}, \bibinfo {author} {\bibfnamefont {J.-n.}\ \bibnamefont
  {Teramae}}, \bibinfo {author} {\bibfnamefont {V.~I.}\ \bibnamefont
  {Nekorkin}}, \ and\ \bibinfo {author} {\bibfnamefont {T.}~\bibnamefont
  {Fukai}},\ }\href {\doibase 10.1371/journal.pone.0094292} {\bibfield
  {journal} {\bibinfo  {journal} {PLoS One}\ }\textbf {\bibinfo {volume} {9}},\
  \bibinfo {pages} {e94292} (\bibinfo {year} {2014})}\BibitemShut {NoStop}%
\bibitem [{\citenamefont {Gisiger}\ and\ \citenamefont
  {Boukadoum}(2011)}]{Gisiger2011}%
  \BibitemOpen
  \bibfield  {author} {\bibinfo {author} {\bibfnamefont {T.}~\bibnamefont
  {Gisiger}}\ and\ \bibinfo {author} {\bibfnamefont {M.}~\bibnamefont
  {Boukadoum}},\ }\href {\doibase 10.3389/fncom.2011.00001} {\bibfield
  {journal} {\bibinfo  {journal} {Front. Comput. Neurosci.}\ }\textbf {\bibinfo
  {volume} {5}},\ \bibinfo {pages} {1} (\bibinfo {year} {2011})}\BibitemShut
  {NoStop}%
\bibitem [{\citenamefont {Johnson}\ \emph {et~al.}(2013)\citenamefont
  {Johnson}, \citenamefont {Marro},\ and\ \citenamefont
  {Torres}}]{Johnson2013}%
  \BibitemOpen
  \bibfield  {author} {\bibinfo {author} {\bibfnamefont {S.}~\bibnamefont
  {Johnson}}, \bibinfo {author} {\bibfnamefont {J.}~\bibnamefont {Marro}}, \
  and\ \bibinfo {author} {\bibfnamefont {J.~J.}\ \bibnamefont {Torres}},\
  }\href {\doibase 10.1371/journal.pone.0050276} {\bibfield  {journal}
  {\bibinfo  {journal} {PLoS One}\ }\textbf {\bibinfo {volume} {8}},\ \bibinfo
  {pages} {e50276} (\bibinfo {year} {2013})}\BibitemShut {NoStop}%
\bibitem [{\citenamefont {Mi{\v{s}}i{\'{c}}}\ \emph {et~al.}(2015)\citenamefont
  {Mi{\v{s}}i{\'{c}}}, \citenamefont {Betzel}, \citenamefont {Nematzadeh},
  \citenamefont {Go{\~{n}}i}, \citenamefont {Griffa}, \citenamefont {Hagmann},
  \citenamefont {Flammini}, \citenamefont {Ahn},\ and\ \citenamefont
  {Sporns}}]{Misic2015}%
  \BibitemOpen
  \bibfield  {author} {\bibinfo {author} {\bibfnamefont {B.}~\bibnamefont
  {Mi{\v{s}}i{\'{c}}}}, \bibinfo {author} {\bibfnamefont {R.~F.}\ \bibnamefont
  {Betzel}}, \bibinfo {author} {\bibfnamefont {A.}~\bibnamefont {Nematzadeh}},
  \bibinfo {author} {\bibfnamefont {J.}~\bibnamefont {Go{\~{n}}i}}, \bibinfo
  {author} {\bibfnamefont {A.}~\bibnamefont {Griffa}}, \bibinfo {author}
  {\bibfnamefont {P.}~\bibnamefont {Hagmann}}, \bibinfo {author} {\bibfnamefont
  {A.}~\bibnamefont {Flammini}}, \bibinfo {author} {\bibfnamefont {Y.-Y.}\
  \bibnamefont {Ahn}}, \ and\ \bibinfo {author} {\bibfnamefont
  {O.}~\bibnamefont {Sporns}},\ }\href {\doibase 10.1016/j.neuron.2015.05.035}
  {\bibfield  {journal} {\bibinfo  {journal} {Neuron}\ }\textbf {\bibinfo
  {volume} {86}},\ \bibinfo {pages} {1518} (\bibinfo {year}
  {2015})}\BibitemShut {NoStop}%
\bibitem [{\citenamefont {Newman}(2003)}]{Newman2003a}%
  \BibitemOpen
  \bibfield  {author} {\bibinfo {author} {\bibfnamefont {M.~E.~J.}\
  \bibnamefont {Newman}},\ }\href {\doibase 10.1137/S003614450342480}
  {\bibfield  {journal} {\bibinfo  {journal} {SIAM Rev.}\ }\textbf {\bibinfo
  {volume} {45}},\ \bibinfo {pages} {167} (\bibinfo {year} {2003})}\BibitemShut
  {NoStop}%
\bibitem [{\citenamefont {Boccaletti}\ \emph {et~al.}(2006)\citenamefont
  {Boccaletti}, \citenamefont {Latora}, \citenamefont {Moreno}, \citenamefont
  {Chavez},\ and\ \citenamefont {Hwang}}]{Boccaletti2006}%
  \BibitemOpen
  \bibfield  {author} {\bibinfo {author} {\bibfnamefont {S.}~\bibnamefont
  {Boccaletti}}, \bibinfo {author} {\bibfnamefont {V.}~\bibnamefont {Latora}},
  \bibinfo {author} {\bibfnamefont {Y.}~\bibnamefont {Moreno}}, \bibinfo
  {author} {\bibfnamefont {M.}~\bibnamefont {Chavez}}, \ and\ \bibinfo {author}
  {\bibfnamefont {D.}~\bibnamefont {Hwang}},\ }\href {\doibase
  10.1016/j.physrep.2005.10.009} {\bibfield  {journal} {\bibinfo  {journal}
  {Phys. Rep.}\ }\textbf {\bibinfo {volume} {424}},\ \bibinfo {pages} {175}
  (\bibinfo {year} {2006})}\BibitemShut {NoStop}%
\bibitem [{\citenamefont {Pastor-Satorras}\ \emph {et~al.}(2015)\citenamefont
  {Pastor-Satorras}, \citenamefont {Castellano}, \citenamefont {{Van
  Mieghem}},\ and\ \citenamefont {Vespignani}}]{Pastor-Satorras2015}%
  \BibitemOpen
  \bibfield  {author} {\bibinfo {author} {\bibfnamefont {R.}~\bibnamefont
  {Pastor-Satorras}}, \bibinfo {author} {\bibfnamefont {C.}~\bibnamefont
  {Castellano}}, \bibinfo {author} {\bibfnamefont {P.}~\bibnamefont {{Van
  Mieghem}}}, \ and\ \bibinfo {author} {\bibfnamefont {A.}~\bibnamefont
  {Vespignani}},\ }\href {\doibase 10.1103/RevModPhys.87.925} {\bibfield
  {journal} {\bibinfo  {journal} {Rev. Mod. Phys.}\ }\textbf {\bibinfo {volume}
  {87}},\ \bibinfo {pages} {925} (\bibinfo {year} {2015})}\BibitemShut
  {NoStop}%
\bibitem [{\citenamefont {Strogatz}(2001)}]{Strogatz2001}%
  \BibitemOpen
  \bibfield  {author} {\bibinfo {author} {\bibfnamefont {S.~H.}\ \bibnamefont
  {Strogatz}},\ }\href {\doibase 10.1038/35065725} {\bibfield  {journal}
  {\bibinfo  {journal} {Nature}\ }\textbf {\bibinfo {volume} {410}},\ \bibinfo
  {pages} {268} (\bibinfo {year} {2001})}\BibitemShut {NoStop}%
\bibitem [{\citenamefont {Chung}\ \emph {et~al.}(2014)\citenamefont {Chung},
  \citenamefont {Baek}, \citenamefont {Kim}, \citenamefont {Ha},\ and\
  \citenamefont {Jeong}}]{Chung2014}%
  \BibitemOpen
  \bibfield  {author} {\bibinfo {author} {\bibfnamefont {K.}~\bibnamefont
  {Chung}}, \bibinfo {author} {\bibfnamefont {Y.}~\bibnamefont {Baek}},
  \bibinfo {author} {\bibfnamefont {D.}~\bibnamefont {Kim}}, \bibinfo {author}
  {\bibfnamefont {M.}~\bibnamefont {Ha}}, \ and\ \bibinfo {author}
  {\bibfnamefont {H.}~\bibnamefont {Jeong}},\ }\href {\doibase
  10.1103/PhysRevE.89.052811} {\bibfield  {journal} {\bibinfo  {journal} {Phys.
  Rev. E}\ }\textbf {\bibinfo {volume} {89}},\ \bibinfo {pages} {052811}
  (\bibinfo {year} {2014})}\BibitemShut {NoStop}%
\bibitem [{\citenamefont {Onnela}\ \emph {et~al.}(2007)\citenamefont {Onnela},
  \citenamefont {Saram{\"{a}}ki}, \citenamefont {Hyv{\"{o}}nen}, \citenamefont
  {Szab{\'{o}}}, \citenamefont {Lazer}, \citenamefont {Kaski}, \citenamefont
  {Kert{\'{e}}sz},\ and\ \citenamefont {Barab{\'{a}}si}}]{Onnela2007}%
  \BibitemOpen
  \bibfield  {author} {\bibinfo {author} {\bibfnamefont {J.-P.}\ \bibnamefont
  {Onnela}}, \bibinfo {author} {\bibfnamefont {J.}~\bibnamefont
  {Saram{\"{a}}ki}}, \bibinfo {author} {\bibfnamefont {J.}~\bibnamefont
  {Hyv{\"{o}}nen}}, \bibinfo {author} {\bibfnamefont {G.}~\bibnamefont
  {Szab{\'{o}}}}, \bibinfo {author} {\bibfnamefont {D.}~\bibnamefont {Lazer}},
  \bibinfo {author} {\bibfnamefont {K.}~\bibnamefont {Kaski}}, \bibinfo
  {author} {\bibfnamefont {J.}~\bibnamefont {Kert{\'{e}}sz}}, \ and\ \bibinfo
  {author} {\bibfnamefont {A.-L.}\ \bibnamefont {Barab{\'{a}}si}},\ }\href
  {\doibase 10.1073/pnas.0610245104} {\bibfield  {journal} {\bibinfo  {journal}
  {Proc. Natl. Acad. Sci. U. S. A.}\ }\textbf {\bibinfo {volume} {104}},\
  \bibinfo {pages} {7332} (\bibinfo {year} {2007})}\BibitemShut {NoStop}%
\bibitem [{\citenamefont {Nematzadeh}\ \emph {et~al.}(2014)\citenamefont
  {Nematzadeh}, \citenamefont {Ferrara}, \citenamefont {Flammini},\ and\
  \citenamefont {Ahn}}]{Nematzadeh2014a}%
  \BibitemOpen
  \bibfield  {author} {\bibinfo {author} {\bibfnamefont {A.}~\bibnamefont
  {Nematzadeh}}, \bibinfo {author} {\bibfnamefont {E.}~\bibnamefont {Ferrara}},
  \bibinfo {author} {\bibfnamefont {A.}~\bibnamefont {Flammini}}, \ and\
  \bibinfo {author} {\bibfnamefont {Y.~Y.}\ \bibnamefont {Ahn}},\ }\href
  {\doibase 10.1103/PhysRevLett.113.088701} {\bibfield  {journal} {\bibinfo
  {journal} {Phys. Rev. Lett.}\ }\textbf {\bibinfo {volume} {113}},\ \bibinfo
  {pages} {1} (\bibinfo {year} {2014})}\BibitemShut {NoStop}%
\bibitem [{\citenamefont {Enel}\ \emph {et~al.}(2016)\citenamefont {Enel},
  \citenamefont {Procyk}, \citenamefont {Quilodran},\ and\ \citenamefont
  {Dominey}}]{Enel2016}%
  \BibitemOpen
  \bibfield  {author} {\bibinfo {author} {\bibfnamefont {P.}~\bibnamefont
  {Enel}}, \bibinfo {author} {\bibfnamefont {E.}~\bibnamefont {Procyk}},
  \bibinfo {author} {\bibfnamefont {R.}~\bibnamefont {Quilodran}}, \ and\
  \bibinfo {author} {\bibfnamefont {P.~F.}\ \bibnamefont {Dominey}},\ }\href
  {\doibase 10.1371/journal.pcbi.1004967} {\bibfield  {journal} {\bibinfo
  {journal} {PLOS Comput. Biol.}\ }\textbf {\bibinfo {volume} {12}},\ \bibinfo
  {pages} {e1004967} (\bibinfo {year} {2016})}\BibitemShut {NoStop}%
\bibitem [{\citenamefont {Soriano}\ \emph {et~al.}(2015)\citenamefont
  {Soriano}, \citenamefont {Brunner}, \citenamefont {Escalona-Moran},
  \citenamefont {Mirasso},\ and\ \citenamefont {Fischer}}]{Soriano2015}%
  \BibitemOpen
  \bibfield  {author} {\bibinfo {author} {\bibfnamefont {M.~C.}\ \bibnamefont
  {Soriano}}, \bibinfo {author} {\bibfnamefont {D.}~\bibnamefont {Brunner}},
  \bibinfo {author} {\bibfnamefont {M.}~\bibnamefont {Escalona-Moran}},
  \bibinfo {author} {\bibfnamefont {C.~R.}\ \bibnamefont {Mirasso}}, \ and\
  \bibinfo {author} {\bibfnamefont {I.}~\bibnamefont {Fischer}},\ }\href
  {\doibase 10.3389/fncom.2015.00068} {\bibfield  {journal} {\bibinfo
  {journal} {Front. Comput. Neurosci.}\ }\textbf {\bibinfo {volume} {9}},\
  \bibinfo {pages} {1} (\bibinfo {year} {2015})}\BibitemShut {NoStop}%
\bibitem [{\citenamefont {Yamazaki}\ and\ \citenamefont
  {Tanaka}(2007)}]{Yamazaki2007}%
  \BibitemOpen
  \bibfield  {author} {\bibinfo {author} {\bibfnamefont {T.}~\bibnamefont
  {Yamazaki}}\ and\ \bibinfo {author} {\bibfnamefont {S.}~\bibnamefont
  {Tanaka}},\ }\href {\doibase 10.1016/j.neunet.2007.04.004} {\bibfield
  {journal} {\bibinfo  {journal} {Neural Networks}\ }\textbf {\bibinfo {volume}
  {20}},\ \bibinfo {pages} {290} (\bibinfo {year} {2007})}\BibitemShut
  {NoStop}%
\bibitem [{\citenamefont {Lukosevicius}\ and\ \citenamefont
  {Jaeger}(2009)}]{Lukosevicius2009}%
  \BibitemOpen
  \bibfield  {author} {\bibinfo {author} {\bibfnamefont {M.}~\bibnamefont
  {Lukosevicius}}\ and\ \bibinfo {author} {\bibfnamefont {H.}~\bibnamefont
  {Jaeger}},\ }\href {\doibase 10.1016/j.cosrev.2009.03.005} {\bibfield
  {journal} {\bibinfo  {journal} {Comput. Sci. Rev.}\ }\textbf {\bibinfo
  {volume} {3}},\ \bibinfo {pages} {127} (\bibinfo {year} {2009})}\BibitemShut
  {NoStop}%
\bibitem [{\citenamefont {Jaeger}\ and\ \citenamefont
  {Hass}(2004)}]{Jaeger2004}%
  \BibitemOpen
  \bibfield  {author} {\bibinfo {author} {\bibfnamefont {H.}~\bibnamefont
  {Jaeger}}\ and\ \bibinfo {author} {\bibfnamefont {H.}~\bibnamefont {Hass}},\
  }\href {\doibase 10.1126/science.1091277} {\bibfield  {journal} {\bibinfo
  {journal} {Science}\ }\textbf {\bibinfo {volume} {304}},\ \bibinfo {pages}
  {78} (\bibinfo {year} {2004})}\BibitemShut {NoStop}%
\bibitem [{\citenamefont {Maass}\ \emph {et~al.}(2002)\citenamefont {Maass},
  \citenamefont {Natschlager},\ and\ \citenamefont {Markram}}]{Maass2002a}%
  \BibitemOpen
  \bibfield  {author} {\bibinfo {author} {\bibfnamefont {W.}~\bibnamefont
  {Maass}}, \bibinfo {author} {\bibfnamefont {T.}~\bibnamefont {Natschlager}},
  \ and\ \bibinfo {author} {\bibfnamefont {H.}~\bibnamefont {Markram}},\ }\href
  {\doibase 10.1162/089976602760407955} {\bibfield  {journal} {\bibinfo
  {journal} {Neural Comput.}\ }\textbf {\bibinfo {volume} {14}},\ \bibinfo
  {pages} {2531} (\bibinfo {year} {2002})}\BibitemShut {NoStop}%
\bibitem [{\citenamefont {Holzmann}\ and\ \citenamefont
  {Hauser}(2010)}]{Holzmann2010}%
  \BibitemOpen
  \bibfield  {author} {\bibinfo {author} {\bibfnamefont {G.}~\bibnamefont
  {Holzmann}}\ and\ \bibinfo {author} {\bibfnamefont {H.}~\bibnamefont
  {Hauser}},\ }\href {\doibase 10.1016/j.neunet.2009.07.004} {\bibfield
  {journal} {\bibinfo  {journal} {Neural Networks}\ }\textbf {\bibinfo {volume}
  {23}},\ \bibinfo {pages} {244} (\bibinfo {year} {2010})}\BibitemShut
  {NoStop}%
\bibitem [{\citenamefont {Jaeger}(2012)}]{Jaeger2012}%
  \BibitemOpen
  \bibfield  {author} {\bibinfo {author} {\bibfnamefont {H.}~\bibnamefont
  {Jaeger}},\ }\href@noop {} {\bibfield  {journal} {\bibinfo  {journal} {Tech.
  report, Jacobs Univ.}\ ,\ \bibinfo {pages} {1}} (\bibinfo {year}
  {2012})}\BibitemShut {NoStop}%
\bibitem [{\citenamefont {Triefenbach}\ \emph {et~al.}(2010)\citenamefont
  {Triefenbach}, \citenamefont {Jalalvand}, \citenamefont {Schrauwen},\ and\
  \citenamefont {Martens}}]{Triefenbach2010}%
  \BibitemOpen
  \bibfield  {author} {\bibinfo {author} {\bibfnamefont {F.}~\bibnamefont
  {Triefenbach}}, \bibinfo {author} {\bibfnamefont {A.}~\bibnamefont
  {Jalalvand}}, \bibinfo {author} {\bibfnamefont {B.}~\bibnamefont
  {Schrauwen}}, \ and\ \bibinfo {author} {\bibfnamefont {J.-P.}\ \bibnamefont
  {Martens}},\ }\href@noop {} {\bibfield  {journal} {\bibinfo  {journal} {Adv.
  Neural Inf. Process. Syst. 23}\ }\textbf {\bibinfo {volume} {23}},\ \bibinfo
  {pages} {1} (\bibinfo {year} {2010})}\BibitemShut {NoStop}%
\bibitem [{\citenamefont {Jalalvand}\ \emph {et~al.}(2016)\citenamefont
  {Jalalvand}, \citenamefont {{De Neve}}, \citenamefont {{Van de Walle}},\ and\
  \citenamefont {Martens}}]{Jalalvand2016}%
  \BibitemOpen
  \bibfield  {author} {\bibinfo {author} {\bibfnamefont {A.}~\bibnamefont
  {Jalalvand}}, \bibinfo {author} {\bibfnamefont {W.}~\bibnamefont {{De
  Neve}}}, \bibinfo {author} {\bibfnamefont {R.}~\bibnamefont {{Van de
  Walle}}}, \ and\ \bibinfo {author} {\bibfnamefont {J.-p.}\ \bibnamefont
  {Martens}},\ }in\ \href {\doibase 10.1109/IJCNN.2016.7727398} {\emph
  {\bibinfo {booktitle} {2016 Int. Jt. Conf. Neural Networks}}}\ (\bibinfo
  {publisher} {IEEE},\ \bibinfo {year} {2016})\ pp.\ \bibinfo {pages}
  {1666--1672}\BibitemShut {NoStop}%
\bibitem [{\citenamefont {Deng}\ \emph {et~al.}(2016)\citenamefont {Deng},
  \citenamefont {Mao},\ and\ \citenamefont {Chen}}]{Deng2016}%
  \BibitemOpen
  \bibfield  {author} {\bibinfo {author} {\bibfnamefont {Z.}~\bibnamefont
  {Deng}}, \bibinfo {author} {\bibfnamefont {C.}~\bibnamefont {Mao}}, \ and\
  \bibinfo {author} {\bibfnamefont {X.}~\bibnamefont {Chen}},\ }in\ \href
  {\doibase 10.1109/IJCNN.2016.7727351} {\emph {\bibinfo {booktitle} {Int. Jt.
  Conf. Neural Networks}}}\ (\bibinfo {year} {2016})\ pp.\ \bibinfo {pages}
  {1325--1332}\BibitemShut {NoStop}%
\bibitem [{\citenamefont {Souahlia}\ \emph {et~al.}(2016)\citenamefont
  {Souahlia}, \citenamefont {Belatreche}, \citenamefont {Benyettou},\ and\
  \citenamefont {Curran}}]{Souahlia2016}%
  \BibitemOpen
  \bibfield  {author} {\bibinfo {author} {\bibfnamefont {A.}~\bibnamefont
  {Souahlia}}, \bibinfo {author} {\bibfnamefont {A.}~\bibnamefont
  {Belatreche}}, \bibinfo {author} {\bibfnamefont {A.}~\bibnamefont
  {Benyettou}}, \ and\ \bibinfo {author} {\bibfnamefont {K.}~\bibnamefont
  {Curran}},\ }in\ \href {\doibase 10.1109/IJCNN.2016.7727326} {\emph {\bibinfo
  {booktitle} {2016 Int. Jt. Conf. Neural Networks}}}\ (\bibinfo  {publisher}
  {IEEE},\ \bibinfo {year} {2016})\ pp.\ \bibinfo {pages}
  {1143--1150}\BibitemShut {NoStop}%
\bibitem [{\citenamefont {R{\"{o}}ssert}\ \emph {et~al.}(2015)\citenamefont
  {R{\"{o}}ssert}, \citenamefont {Dean},\ and\ \citenamefont
  {Porrill}}]{Rossert2015}%
  \BibitemOpen
  \bibfield  {author} {\bibinfo {author} {\bibfnamefont {C.}~\bibnamefont
  {R{\"{o}}ssert}}, \bibinfo {author} {\bibfnamefont {P.}~\bibnamefont {Dean}},
  \ and\ \bibinfo {author} {\bibfnamefont {J.}~\bibnamefont {Porrill}},\ }\href
  {\doibase 10.1371/journal.pcbi.1004515} {\bibfield  {journal} {\bibinfo
  {journal} {PLoS Comput. Biol.}\ }\textbf {\bibinfo {volume} {11}},\ \bibinfo
  {pages} {1} (\bibinfo {year} {2015})}\BibitemShut {NoStop}%
\bibitem [{\citenamefont {Pascanu}\ and\ \citenamefont
  {Jaeger}(2011)}]{Pascanu2011a}%
  \BibitemOpen
  \bibfield  {author} {\bibinfo {author} {\bibfnamefont {R.}~\bibnamefont
  {Pascanu}}\ and\ \bibinfo {author} {\bibfnamefont {H.}~\bibnamefont
  {Jaeger}},\ }\href {\doibase 10.1016/j.neunet.2010.10.003} {\bibfield
  {journal} {\bibinfo  {journal} {Neural Networks}\ }\textbf {\bibinfo {volume}
  {24}},\ \bibinfo {pages} {199} (\bibinfo {year} {2011})}\BibitemShut
  {NoStop}%
\bibitem [{\citenamefont {Jaeger}(2002)}]{Jaeger2002}%
  \BibitemOpen
  \bibfield  {author} {\bibinfo {author} {\bibfnamefont {H.}~\bibnamefont
  {Jaeger}},\ }\href@noop {} {\bibfield  {journal} {\bibinfo  {journal} {GMD
  Rep. 152}\ ,\ \bibinfo {pages} {60}} (\bibinfo {year} {2002})}\BibitemShut
  {NoStop}%
\bibitem [{\citenamefont {Rodan}\ and\ \citenamefont
  {Tiňo}(2011)}]{Rodan2011}%
  \BibitemOpen
  \bibfield  {author} {\bibinfo {author} {\bibfnamefont {A.}~\bibnamefont
  {Rodan}}\ and\ \bibinfo {author} {\bibfnamefont {P.}~\bibnamefont {Tiňo}},\
  }\href {\doibase 10.1109/TNN.2010.2089641} {\bibfield  {journal} {\bibinfo
  {journal} {IEEE Trans. Neural Networks}\ }\textbf {\bibinfo {volume} {22}},\
  \bibinfo {pages} {131} (\bibinfo {year} {2011})}\BibitemShut {NoStop}%
\bibitem [{\citenamefont {Farkas}\ \emph {et~al.}(2016)\citenamefont {Farkas},
  \citenamefont {Bosak},\ and\ \citenamefont {Gergel}}]{Farkas2016}%
  \BibitemOpen
  \bibfield  {author} {\bibinfo {author} {\bibfnamefont {I.}~\bibnamefont
  {Farkas}}, \bibinfo {author} {\bibfnamefont {R.}~\bibnamefont {Bosak}}, \
  and\ \bibinfo {author} {\bibfnamefont {P.}~\bibnamefont {Gergel}},\ }\href
  {\doibase 10.1016/j.neunet.2016.07.012} {\bibfield  {journal} {\bibinfo
  {journal} {Neural Networks}\ }\textbf {\bibinfo {volume} {83}},\ \bibinfo
  {pages} {109} (\bibinfo {year} {2016})}\BibitemShut {NoStop}%
\bibitem [{\citenamefont {Verstraeten}\ \emph {et~al.}(2007)\citenamefont
  {Verstraeten}, \citenamefont {Schrauwen}, \citenamefont {D'Haene},\ and\
  \citenamefont {Stroobandt}}]{Verstraeten2007}%
  \BibitemOpen
  \bibfield  {author} {\bibinfo {author} {\bibfnamefont {D.}~\bibnamefont
  {Verstraeten}}, \bibinfo {author} {\bibfnamefont {B.}~\bibnamefont
  {Schrauwen}}, \bibinfo {author} {\bibfnamefont {M.}~\bibnamefont {D'Haene}},
  \ and\ \bibinfo {author} {\bibfnamefont {D.}~\bibnamefont {Stroobandt}},\
  }\href {\doibase 10.1016/j.neunet.2007.04.003} {\bibfield  {journal}
  {\bibinfo  {journal} {Neural Networks}\ }\textbf {\bibinfo {volume} {20}},\
  \bibinfo {pages} {391} (\bibinfo {year} {2007})}\BibitemShut {NoStop}%
\bibitem [{\citenamefont {Ju}\ \emph {et~al.}(2013)\citenamefont {Ju},
  \citenamefont {Xu}, \citenamefont {Chong},\ and\ \citenamefont
  {VanDongen}}]{Ju2013}%
  \BibitemOpen
  \bibfield  {author} {\bibinfo {author} {\bibfnamefont {H.}~\bibnamefont
  {Ju}}, \bibinfo {author} {\bibfnamefont {J.~X.}\ \bibnamefont {Xu}}, \bibinfo
  {author} {\bibfnamefont {E.}~\bibnamefont {Chong}}, \ and\ \bibinfo {author}
  {\bibfnamefont {A.~M.~J.}\ \bibnamefont {VanDongen}},\ }\href {\doibase
  10.1016/j.neunet.2012.11.003} {\bibfield  {journal} {\bibinfo  {journal}
  {Neural Networks}\ }\textbf {\bibinfo {volume} {38}},\ \bibinfo {pages} {39}
  (\bibinfo {year} {2013})}\BibitemShut {NoStop}%
\bibitem [{\citenamefont {Deng}\ and\ \citenamefont {Zhang}(2007)}]{Deng2007}%
  \BibitemOpen
  \bibfield  {author} {\bibinfo {author} {\bibfnamefont {Z.}~\bibnamefont
  {Deng}}\ and\ \bibinfo {author} {\bibfnamefont {Y.}~\bibnamefont {Zhang}},\
  }\href@noop {} {\bibfield  {journal} {\bibinfo  {journal} {IEEE Trans. Neural
  Networks}\ }\textbf {\bibinfo {volume} {18}},\ \bibinfo {pages} {1364}
  (\bibinfo {year} {2007})}\BibitemShut {NoStop}%
\bibitem [{\citenamefont {Li}\ \emph {et~al.}(2015)\citenamefont {Li},
  \citenamefont {Zhong}, \citenamefont {Xue},\ and\ \citenamefont
  {Zhang}}]{Li2015}%
  \BibitemOpen
  \bibfield  {author} {\bibinfo {author} {\bibfnamefont {X.}~\bibnamefont
  {Li}}, \bibinfo {author} {\bibfnamefont {L.}~\bibnamefont {Zhong}}, \bibinfo
  {author} {\bibfnamefont {F.}~\bibnamefont {Xue}}, \ and\ \bibinfo {author}
  {\bibfnamefont {A.}~\bibnamefont {Zhang}},\ }\href {\doibase
  10.1371/journal.pone.0120750} {\bibfield  {journal} {\bibinfo  {journal}
  {PLoS One}\ }\textbf {\bibinfo {volume} {10}},\ \bibinfo {pages} {1}
  (\bibinfo {year} {2015})}\BibitemShut {NoStop}%
\bibitem [{\citenamefont {Rad}\ \emph {et~al.}(2008)\citenamefont {Rad},
  \citenamefont {Jalili},\ and\ \citenamefont {Hasler}}]{Rad2008}%
  \BibitemOpen
  \bibfield  {author} {\bibinfo {author} {\bibfnamefont {A.~A.}\ \bibnamefont
  {Rad}}, \bibinfo {author} {\bibfnamefont {M.}~\bibnamefont {Jalili}}, \ and\
  \bibinfo {author} {\bibfnamefont {M.}~\bibnamefont {Hasler}},\ }\href@noop {}
  {\bibfield  {journal} {\bibinfo  {journal} {IEEE Int. Symp. Circuits Syst.}\
  ,\ \bibinfo {pages} {868}} (\bibinfo {year} {2008})}\BibitemShut {NoStop}%
\bibitem [{\citenamefont {Leskovec}\ \emph {et~al.}(2010)\citenamefont
  {Leskovec}, \citenamefont {Chakrabarti}, \citenamefont {Kleinberg},
  \citenamefont {Faloutsos},\ and\ \citenamefont {Ghahramani}}]{Leskovec2010}%
  \BibitemOpen
  \bibfield  {author} {\bibinfo {author} {\bibfnamefont {J.}~\bibnamefont
  {Leskovec}}, \bibinfo {author} {\bibfnamefont {D.}~\bibnamefont
  {Chakrabarti}}, \bibinfo {author} {\bibfnamefont {J.}~\bibnamefont
  {Kleinberg}}, \bibinfo {author} {\bibfnamefont {C.}~\bibnamefont
  {Faloutsos}}, \ and\ \bibinfo {author} {\bibfnamefont {Z.}~\bibnamefont
  {Ghahramani}},\ }\href {\doibase 10.1145/1756006.1756039} {\bibfield
  {journal} {\bibinfo  {journal} {J. Mach. Learn. Res.}\ }\textbf {\bibinfo
  {volume} {11}},\ \bibinfo {pages} {985} (\bibinfo {year} {2010})}\BibitemShut
  {NoStop}%
\bibitem [{\citenamefont {Xue}\ \emph {et~al.}(2007)\citenamefont {Xue},
  \citenamefont {Yang},\ and\ \citenamefont {Haykin}}]{Xue2007}%
  \BibitemOpen
  \bibfield  {author} {\bibinfo {author} {\bibfnamefont {Y.}~\bibnamefont
  {Xue}}, \bibinfo {author} {\bibfnamefont {L.}~\bibnamefont {Yang}}, \ and\
  \bibinfo {author} {\bibfnamefont {S.}~\bibnamefont {Haykin}},\ }\href
  {\doibase 10.1016/j.neunet.2007.04.014} {\bibfield  {journal} {\bibinfo
  {journal} {Neural Networks}\ }\textbf {\bibinfo {volume} {20}},\ \bibinfo
  {pages} {365} (\bibinfo {year} {2007})}\BibitemShut {NoStop}%
\bibitem [{\citenamefont {Dambre}\ \emph {et~al.}(2012)\citenamefont {Dambre},
  \citenamefont {Verstraeten}, \citenamefont {Schrauwen},\ and\ \citenamefont
  {Massar}}]{Dambre2012}%
  \BibitemOpen
  \bibfield  {author} {\bibinfo {author} {\bibfnamefont {J.}~\bibnamefont
  {Dambre}}, \bibinfo {author} {\bibfnamefont {D.}~\bibnamefont {Verstraeten}},
  \bibinfo {author} {\bibfnamefont {B.}~\bibnamefont {Schrauwen}}, \ and\
  \bibinfo {author} {\bibfnamefont {S.}~\bibnamefont {Massar}},\ }\href
  {\doibase 10.1038/srep00514} {\bibfield  {journal} {\bibinfo  {journal} {Sci.
  Rep.}\ }\textbf {\bibinfo {volume} {2}},\ \bibinfo {pages} {514} (\bibinfo
  {year} {2012})}\BibitemShut {NoStop}%
\bibitem [{\citenamefont {Verstraeten}\ \emph {et~al.}(2010)\citenamefont
  {Verstraeten}, \citenamefont {Dambre}, \citenamefont {Dutoit},\ and\
  \citenamefont {Schrauwen}}]{Verstraeten2010}%
  \BibitemOpen
  \bibfield  {author} {\bibinfo {author} {\bibfnamefont {D.}~\bibnamefont
  {Verstraeten}}, \bibinfo {author} {\bibfnamefont {J.}~\bibnamefont {Dambre}},
  \bibinfo {author} {\bibfnamefont {X.}~\bibnamefont {Dutoit}}, \ and\ \bibinfo
  {author} {\bibfnamefont {B.}~\bibnamefont {Schrauwen}},\ }in\ \href {\doibase
  10.1109/IJCNN.2010.5596492} {\emph {\bibinfo {booktitle} {2010 Int. Jt. Conf.
  Neural Networks}}}\ (\bibinfo  {publisher} {IEEE},\ \bibinfo {year} {2010})\
  pp.\ \bibinfo {pages} {1--8}\BibitemShut {NoStop}%
\bibitem [{\citenamefont {Luko{\v{s}}evi{\v{c}}ius}(2012)}]{Lukosevicius2012a}%
  \BibitemOpen
  \bibfield  {author} {\bibinfo {author} {\bibfnamefont {M.}~\bibnamefont
  {Luko{\v{s}}evi{\v{c}}ius}},\ }\href@noop {} {\bibfield  {journal} {\bibinfo
  {journal} {Neural Networks: Tricks of the Trade, Reloaded}\ ,\ \bibinfo
  {pages} {659}} (\bibinfo {year} {2012})}\BibitemShut {NoStop}%
\bibitem [{\citenamefont {Garcia}\ \emph {et~al.}(2014)\citenamefont {Garcia},
  \citenamefont {Lesne}, \citenamefont {Hilgetag},\ and\ \citenamefont
  {H{\"{u}}tt}}]{Garcia2014}%
  \BibitemOpen
  \bibfield  {author} {\bibinfo {author} {\bibfnamefont {G.~C.}\ \bibnamefont
  {Garcia}}, \bibinfo {author} {\bibfnamefont {A.}~\bibnamefont {Lesne}},
  \bibinfo {author} {\bibfnamefont {C.~C.}\ \bibnamefont {Hilgetag}}, \ and\
  \bibinfo {author} {\bibfnamefont {M.~T.}\ \bibnamefont {H{\"{u}}tt}},\ }\href
  {\doibase 10.1103/PhysRevE.90.052805} {\bibfield  {journal} {\bibinfo
  {journal} {Phys. Rev. E - Stat. Nonlinear, Soft Matter Phys.}\ }\textbf
  {\bibinfo {volume} {90}},\ \bibinfo {pages} {1} (\bibinfo {year}
  {2014})}\BibitemShut {NoStop}%
\bibitem [{\citenamefont {Bertschinger}\ and\ \citenamefont
  {Natschl{\"{a}}ger}(2004)}]{Bertschinger2004}%
  \BibitemOpen
  \bibfield  {author} {\bibinfo {author} {\bibfnamefont {N.}~\bibnamefont
  {Bertschinger}}\ and\ \bibinfo {author} {\bibfnamefont {T.}~\bibnamefont
  {Natschl{\"{a}}ger}},\ }\href {\doibase 10.1162/089976604323057443}
  {\bibfield  {journal} {\bibinfo  {journal} {Neural Comput.}\ }\textbf
  {\bibinfo {volume} {16}},\ \bibinfo {pages} {1413} (\bibinfo {year}
  {2004})}\BibitemShut {NoStop}%
\bibitem [{\citenamefont {Haeusler}\ and\ \citenamefont
  {Maass}(2007)}]{Haeusler2007}%
  \BibitemOpen
  \bibfield  {author} {\bibinfo {author} {\bibfnamefont {S.}~\bibnamefont
  {Haeusler}}\ and\ \bibinfo {author} {\bibfnamefont {W.}~\bibnamefont
  {Maass}},\ }\href {\doibase 10.1093/cercor/bhj132} {\bibfield  {journal}
  {\bibinfo  {journal} {Cereb. Cortex}\ }\textbf {\bibinfo {volume} {17}},\
  \bibinfo {pages} {149} (\bibinfo {year} {2007})}\BibitemShut {NoStop}%
\bibitem [{\citenamefont {Schrauwen}\ \emph {et~al.}(2009)\citenamefont
  {Schrauwen}, \citenamefont {Buesing},\ and\ \citenamefont
  {Legenstein}}]{Schrauwen2009}%
  \BibitemOpen
  \bibfield  {author} {\bibinfo {author} {\bibfnamefont {B.}~\bibnamefont
  {Schrauwen}}, \bibinfo {author} {\bibfnamefont {L.}~\bibnamefont {Buesing}},
  \ and\ \bibinfo {author} {\bibfnamefont {R.}~\bibnamefont {Legenstein}},\
  }\href@noop {} {\bibfield  {journal} {\bibinfo  {journal} {Adv. Neural Inf.
  Process. Syst.}\ }\textbf {\bibinfo {volume} {21}},\ \bibinfo {pages} {1425}
  (\bibinfo {year} {2009})}\BibitemShut {NoStop}%
\bibitem [{\citenamefont {Hilgetag}\ \emph {et~al.}(2000)\citenamefont
  {Hilgetag}, \citenamefont {Burns}, \citenamefont {O'Neill}, \citenamefont
  {Scannell},\ and\ \citenamefont {Young}}]{Hilgetag2000}%
  \BibitemOpen
  \bibfield  {author} {\bibinfo {author} {\bibfnamefont {C.~C.}\ \bibnamefont
  {Hilgetag}}, \bibinfo {author} {\bibfnamefont {G.~a.}\ \bibnamefont {Burns}},
  \bibinfo {author} {\bibfnamefont {M.~a.}\ \bibnamefont {O'Neill}}, \bibinfo
  {author} {\bibfnamefont {J.~W.}\ \bibnamefont {Scannell}}, \ and\ \bibinfo
  {author} {\bibfnamefont {M.~P.}\ \bibnamefont {Young}},\ }\href {\doibase
  10.1098/rstb.2000.0551} {\bibfield  {journal} {\bibinfo  {journal} {Philos.
  Trans. R. Soc. Lond. B. Biol. Sci.}\ }\textbf {\bibinfo {volume} {355}},\
  \bibinfo {pages} {91} (\bibinfo {year} {2000})}\BibitemShut {NoStop}%
\bibitem [{\citenamefont {Meunier}\ \emph {et~al.}(2010)\citenamefont
  {Meunier}, \citenamefont {Lambiotte},\ and\ \citenamefont
  {Bullmore}}]{Meunier2010}%
  \BibitemOpen
  \bibfield  {author} {\bibinfo {author} {\bibfnamefont {D.}~\bibnamefont
  {Meunier}}, \bibinfo {author} {\bibfnamefont {R.}~\bibnamefont {Lambiotte}},
  \ and\ \bibinfo {author} {\bibfnamefont {E.~T.}\ \bibnamefont {Bullmore}},\
  }\href {\doibase 10.3389/fnins.2010.00200} {\bibfield  {journal} {\bibinfo
  {journal} {Front. Neurosci.}\ }\textbf {\bibinfo {volume} {4}},\ \bibinfo
  {pages} {1} (\bibinfo {year} {2010})}\BibitemShut {NoStop}%
\bibitem [{\citenamefont {Sporns}\ \emph {et~al.}(2004)\citenamefont {Sporns},
  \citenamefont {Chialvo}, \citenamefont {Kaiser},\ and\ \citenamefont
  {Hilgetag}}]{Sporns2004}%
  \BibitemOpen
  \bibfield  {author} {\bibinfo {author} {\bibfnamefont {O.}~\bibnamefont
  {Sporns}}, \bibinfo {author} {\bibfnamefont {D.~R.}\ \bibnamefont {Chialvo}},
  \bibinfo {author} {\bibfnamefont {M.}~\bibnamefont {Kaiser}}, \ and\ \bibinfo
  {author} {\bibfnamefont {C.~C.}\ \bibnamefont {Hilgetag}},\ }\href {\doibase
  10.1016/j.tics.2004.07.008} {\bibfield  {journal} {\bibinfo  {journal}
  {Trends Cogn. Sci.}\ }\textbf {\bibinfo {volume} {8}},\ \bibinfo {pages}
  {418} (\bibinfo {year} {2004})}\BibitemShut {NoStop}%
\bibitem [{\citenamefont {Hagmann}\ \emph {et~al.}(2008)\citenamefont
  {Hagmann}, \citenamefont {Cammoun}, \citenamefont {Gigandet}, \citenamefont
  {Meuli}, \citenamefont {Honey}, \citenamefont {{Van Wedeen}},\ and\
  \citenamefont {Sporns}}]{Hagmann2008}%
  \BibitemOpen
  \bibfield  {author} {\bibinfo {author} {\bibfnamefont {P.}~\bibnamefont
  {Hagmann}}, \bibinfo {author} {\bibfnamefont {L.}~\bibnamefont {Cammoun}},
  \bibinfo {author} {\bibfnamefont {X.}~\bibnamefont {Gigandet}}, \bibinfo
  {author} {\bibfnamefont {R.}~\bibnamefont {Meuli}}, \bibinfo {author}
  {\bibfnamefont {C.~J.}\ \bibnamefont {Honey}}, \bibinfo {author}
  {\bibfnamefont {J.}~\bibnamefont {{Van Wedeen}}}, \ and\ \bibinfo {author}
  {\bibfnamefont {O.}~\bibnamefont {Sporns}},\ }\href {\doibase
  10.1371/journal.pbio.0060159} {\bibfield  {journal} {\bibinfo  {journal}
  {PLoS Biol.}\ }\textbf {\bibinfo {volume} {6}},\ \bibinfo {pages} {1479}
  (\bibinfo {year} {2008})}\BibitemShut {NoStop}%
\bibitem [{\citenamefont {Shimono}\ and\ \citenamefont
  {Beggs}(2015)}]{Shimono2015a}%
  \BibitemOpen
  \bibfield  {author} {\bibinfo {author} {\bibfnamefont {M.}~\bibnamefont
  {Shimono}}\ and\ \bibinfo {author} {\bibfnamefont {J.~M.}\ \bibnamefont
  {Beggs}},\ }\href {\doibase 10.1093/cercor/bhu252} {\bibfield  {journal}
  {\bibinfo  {journal} {Cereb. Cortex}\ }\textbf {\bibinfo {volume} {25}},\
  \bibinfo {pages} {3743} (\bibinfo {year} {2015})}\BibitemShut {NoStop}%
\bibitem [{\citenamefont {Buxhoeveden}\ and\ \citenamefont
  {Casanova}(2002)}]{Buxhoeveden2002}%
  \BibitemOpen
  \bibfield  {author} {\bibinfo {author} {\bibfnamefont {D.~P.}\ \bibnamefont
  {Buxhoeveden}}\ and\ \bibinfo {author} {\bibfnamefont {M.~F.}\ \bibnamefont
  {Casanova}},\ }\href {\doibase 10.1093/brain/awf110} {\bibfield  {journal}
  {\bibinfo  {journal} {Brain}\ }\textbf {\bibinfo {volume} {125}},\ \bibinfo
  {pages} {935} (\bibinfo {year} {2002})}\BibitemShut {NoStop}%
\bibitem [{\citenamefont {Simon}(1997)}]{Simon1997}%
  \BibitemOpen
  \bibfield  {author} {\bibinfo {author} {\bibfnamefont {H.~a.}\ \bibnamefont
  {Simon}},\ }\href@noop {} {\emph {\bibinfo {title} {{The sciences of the
  artificial}}}},\ \bibinfo {edition} {3rd}\ ed.\ (\bibinfo  {publisher} {MIT
  Press},\ \bibinfo {address} {Cambridge, Mass.},\ \bibinfo {year} {1997})\ p.\
  \bibinfo {pages} {130}\BibitemShut {NoStop}%
\bibitem [{\citenamefont {Kashtan}\ and\ \citenamefont
  {Alon}(2005)}]{Kashtan2005}%
  \BibitemOpen
  \bibfield  {author} {\bibinfo {author} {\bibfnamefont {N.}~\bibnamefont
  {Kashtan}}\ and\ \bibinfo {author} {\bibfnamefont {U.}~\bibnamefont {Alon}},\
  }\href {\doibase 10.1073/pnas.0503610102} {\bibfield  {journal} {\bibinfo
  {journal} {Proc. Natl. Acad. Sci. U. S. A.}\ }\textbf {\bibinfo {volume}
  {102}},\ \bibinfo {pages} {13773} (\bibinfo {year} {2005})}\BibitemShut
  {NoStop}%
\bibitem [{\citenamefont {Clune}\ \emph {et~al.}(2013)\citenamefont {Clune},
  \citenamefont {Mouret},\ and\ \citenamefont {Lipson}}]{Clune2013}%
  \BibitemOpen
  \bibfield  {author} {\bibinfo {author} {\bibfnamefont {J.}~\bibnamefont
  {Clune}}, \bibinfo {author} {\bibfnamefont {J.-B.}\ \bibnamefont {Mouret}}, \
  and\ \bibinfo {author} {\bibfnamefont {H.}~\bibnamefont {Lipson}},\ }\href
  {\doibase 10.1098/rspb.2012.2863} {\bibfield  {journal} {\bibinfo  {journal}
  {Proc. R. Soc. London B Biol. Sci.}\ }\textbf {\bibinfo {volume} {280}},\
  \bibinfo {pages} {20122863} (\bibinfo {year} {2013})}\BibitemShut {NoStop}%
\bibitem [{\citenamefont {Ellefsen}\ \emph {et~al.}(2015)\citenamefont
  {Ellefsen}, \citenamefont {Mouret},\ and\ \citenamefont
  {Clune}}]{Ellefsen2015}%
  \BibitemOpen
  \bibfield  {author} {\bibinfo {author} {\bibfnamefont {K.~O.}\ \bibnamefont
  {Ellefsen}}, \bibinfo {author} {\bibfnamefont {J.~B.}\ \bibnamefont
  {Mouret}}, \ and\ \bibinfo {author} {\bibfnamefont {J.}~\bibnamefont
  {Clune}},\ }\href {\doibase 10.1371/journal.pcbi.1004128} {\bibfield
  {journal} {\bibinfo  {journal} {PLoS Comput. Biol.}\ }\textbf {\bibinfo
  {volume} {11}},\ \bibinfo {pages} {1} (\bibinfo {year} {2015})}\BibitemShut
  {NoStop}%
\bibitem [{\citenamefont {Kaiser}\ and\ \citenamefont
  {Hilgetag}(2004)}]{Kaiser2004}%
  \BibitemOpen
  \bibfield  {author} {\bibinfo {author} {\bibfnamefont {M.}~\bibnamefont
  {Kaiser}}\ and\ \bibinfo {author} {\bibfnamefont {C.~C.}\ \bibnamefont
  {Hilgetag}},\ }\href {\doibase 10.1103/PhysRevE.69.036103} {\bibfield
  {journal} {\bibinfo  {journal} {Phys. Rev. E}\ }\textbf {\bibinfo {volume}
  {69}},\ \bibinfo {pages} {1} (\bibinfo {year} {2004})}\BibitemShut {NoStop}%
\bibitem [{\citenamefont {Lancichinetti}\ \emph {et~al.}(2008)\citenamefont
  {Lancichinetti}, \citenamefont {Fortunato},\ and\ \citenamefont
  {Radicchi}}]{Lancichinetti2008}%
  \BibitemOpen
  \bibfield  {author} {\bibinfo {author} {\bibfnamefont {A.}~\bibnamefont
  {Lancichinetti}}, \bibinfo {author} {\bibfnamefont {S.}~\bibnamefont
  {Fortunato}}, \ and\ \bibinfo {author} {\bibfnamefont {F.}~\bibnamefont
  {Radicchi}},\ }\href {\doibase 10.1103/PhysRevE.78.046110} {\bibfield
  {journal} {\bibinfo  {journal} {Phys. Rev. E}\ }\textbf {\bibinfo {volume}
  {78}},\ \bibinfo {pages} {1} (\bibinfo {year} {2008})}\BibitemShut {NoStop}%
\bibitem [{\citenamefont {Newman}(210)}]{Newman2010B}%
  \BibitemOpen
  \bibfield  {author} {\bibinfo {author} {\bibfnamefont {M.~E.~J.}\
  \bibnamefont {Newman}},\ }in\ \href@noop {} {\emph {\bibinfo {booktitle}
  {Networks An Introd.}}}\ (\bibinfo  {publisher} {Oxford},\ \bibinfo {year}
  {210})\ Chap.~\bibinfo {chapter} {13}, pp.\ \bibinfo {pages}
  {434--444}\BibitemShut {NoStop}%
\bibitem [{\citenamefont {Azadeh}\ \emph {et~al.}(2018)\citenamefont {Azadeh},
  \citenamefont {Rodriguez}, \citenamefont {Flammini},\ and\ \citenamefont
  {Ahn}}]{azadeh2018}%
  \BibitemOpen
  \bibfield  {author} {\bibinfo {author} {\bibfnamefont {N.}~\bibnamefont
  {Azadeh}}, \bibinfo {author} {\bibfnamefont {N.}~\bibnamefont {Rodriguez}},
  \bibinfo {author} {\bibfnamefont {A.}~\bibnamefont {Flammini}}, \ and\
  \bibinfo {author} {\bibfnamefont {Y.-Y.}\ \bibnamefont {Ahn}},\ }in\
  \href@noop {} {\emph {\bibinfo {booktitle} {Spreading Dyn. Soc. Syst.}}},\
  \bibinfo {editor} {edited by\ \bibinfo {editor} {\bibfnamefont
  {S.}~\bibnamefont {Lehmann}}\ and\ \bibinfo {editor} {\bibfnamefont {Y.-Y.}\
  \bibnamefont {Ahn}}}\ (\bibinfo  {publisher} {Springer},\ \bibinfo {year}
  {2018})\ Chap.~\bibinfo {chapter} {2}\BibitemShut {NoStop}%
\end{thebibliography}%

\newpage

\onecolumngrid
\begin{center}
\textbf{\large Supplemental Materials: Optimal modularity and memory capacity of neural networks}
\end{center}
%%%%%%%%%% Merge with supplemental materials %%%%%%%%%%
%%%%%%%%%% Prefix a "S" to all equations, figures, tables and reset the counter %%%%%%%%%%
\setcounter{equation}{0}
\setcounter{figure}{0}
\setcounter{table}{0}
\makeatletter
\renewcommand{\theequation}{S\arabic{equation}}
\renewcommand{\thefigure}{S\arabic{figure}}
\renewcommand\thetable{S.\arabic{table}}
\renewcommand{\bibnumfmt}[1]{[S#1]}
\renewcommand{\citenumfont}[1]{S#1}

\begin{table}[htb]
\centering
\begin{tabular}{|c|c|c|c|c|c|}
\hline
Technique & Realism & Training & Structural Control & Task Flexibility \\
\hline\hline
ESN & model choice & simple, convex & full & high \\
RNN & model choice & complex, non-convex & low & low \\
LSTM & low & complex, non-convex & low & low \\
\hline
\end{tabular}
\caption{A break-down of the differences between ESNs and a couple other artificial neural network approaches that can be trained on complex tasks. ESNs provide an ideal environment for testing our hypothesis due to their flexibility. Traditional RNNs and LSTMs (long-short term memory neural networks) require more complex training procedures which alter the synaptic weights between neurons resulting in loss of control over network structure impacting modularity and other network properties. Alternatively, reservoirs in ESNs are static allowing us to readily control the properties of the network that we want to investigate. Additionally, ESN reservoirs can be swapped from task to task without modification due to the separate input and read-out layers, whereas RNNs and LSTMs need entirely new networks for each task. Lastly, ESNs and RNNs can be made more biologically realistic by including higher fidelity models for their components, such as spiking neurons or plastic synapses. We opted against this due to the added complexity that comes with continuous-time spiking neurons which require special encoding/decoding rules for task inputs and outputs.}
\label{table:tech}
\end{table}

\begin{figure}[tbhp!]
\centering
\includegraphics[width=0.9\linewidth]{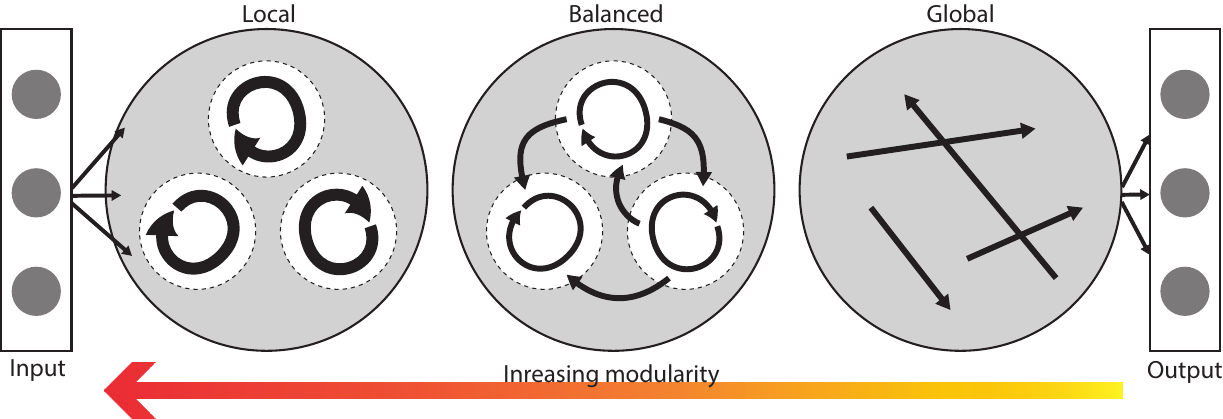}
\caption{The modularity of the reservoir of the ESN can vary from tightly coupled communities to a completely random network. In the ``local'' domain information travels freely within communities but has difficulty escaping and isn't shared between communities. Alternatively, in the ``global'' domain neurons have connections throughout the reservoir and information should flow freely, but due to the lack of local cohesion provided by communities neurons fail to activate and information flow is quenched. The read-out layer of the ESN is connected to all the neurons in the reservoir, however it is only a linear aggregation of the reservoir state. To take advantage of the non-linearity of the reservoir information would need to be maintained and processed within it. We argue that these benefits are best achieved in a balanced region where local cohesion and global connectivity facilitate global information spreading.}
\label{fig:concept_diagram}
\end{figure}

\begin{figure}[tbhp!]
\centering
\includegraphics[width=0.9\linewidth]{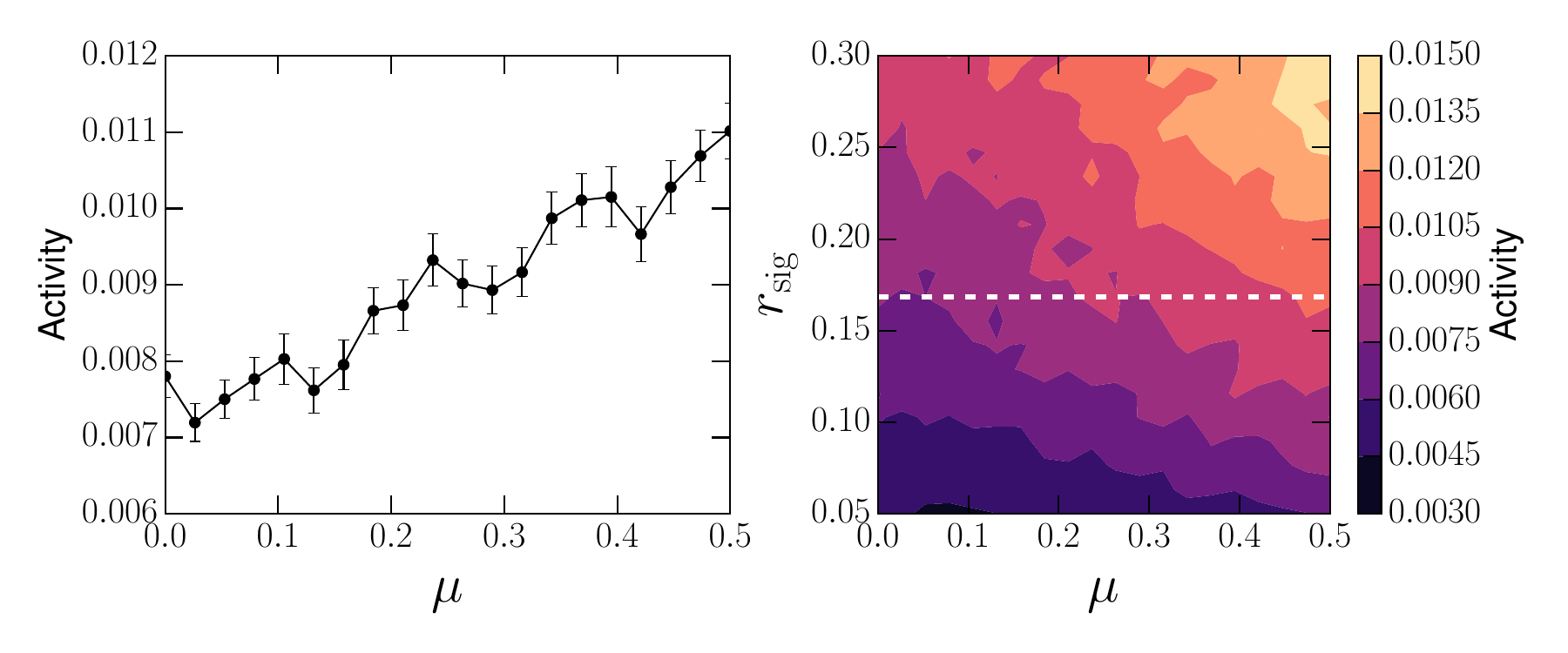}
\caption{The absolute value of the activity response of a linear reservoir. As $\mu$ decreases so does the activity. (left) Shows a slice through the phase diagram at the dashed white line. (right) The activity as $\mu$ and input fraction vary. It uses graphs with $N=500$, $k=6$, a community size of $10$. The spectral radius has been shown to have a large impact on linear and tanh reservoir behavior, so we correct for changes in spectral radius as $\mu$ changes. The reservoir is kept just below the critical point with $W_s=0.9$ to prevent numerical explosions.}
\label{fig:linear_activation_results}
\end{figure}

\begin{figure}[tbhp!]
\centering
\includegraphics[width=0.9\linewidth]{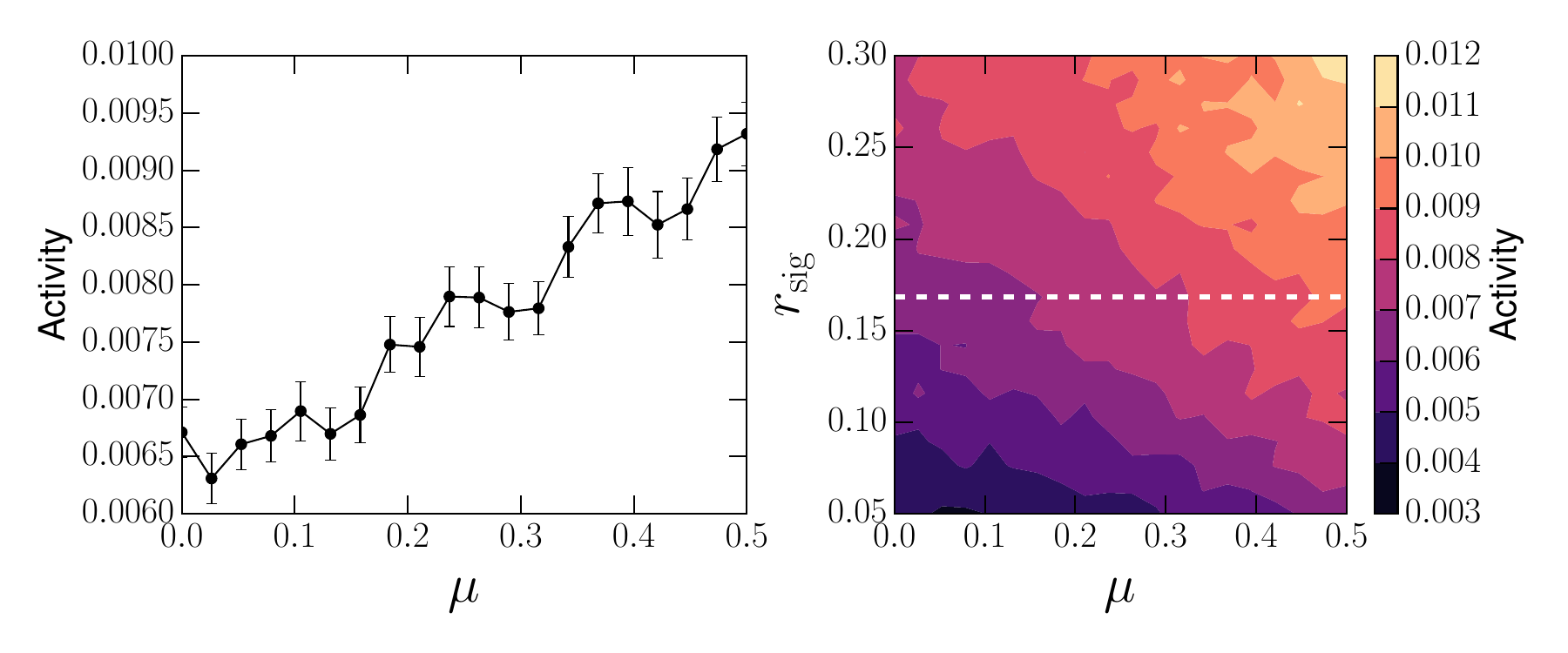}
\caption{The absolute value of the activity response of a tanh reservoir. As $\mu$ decreases so does the activity. (left) Shows a slice through the phase diagram at the dashed white line. (right) The activity as $\mu$ and input fraction vary. It uses graphs with $N=500$, $k=6$, a community size of $10$. The spectral radius has been shown to have a large impact on linear and tanh reservoir behavior, so we correct for changes in spectral radius as $\mu$ changes. The reservoir is kept at the critical point with $W_s=1.0$.}
\label{fig:tanh_activation_results}
\end{figure}

\begin{figure}[tbhp!]
\includegraphics[width=\linewidth]{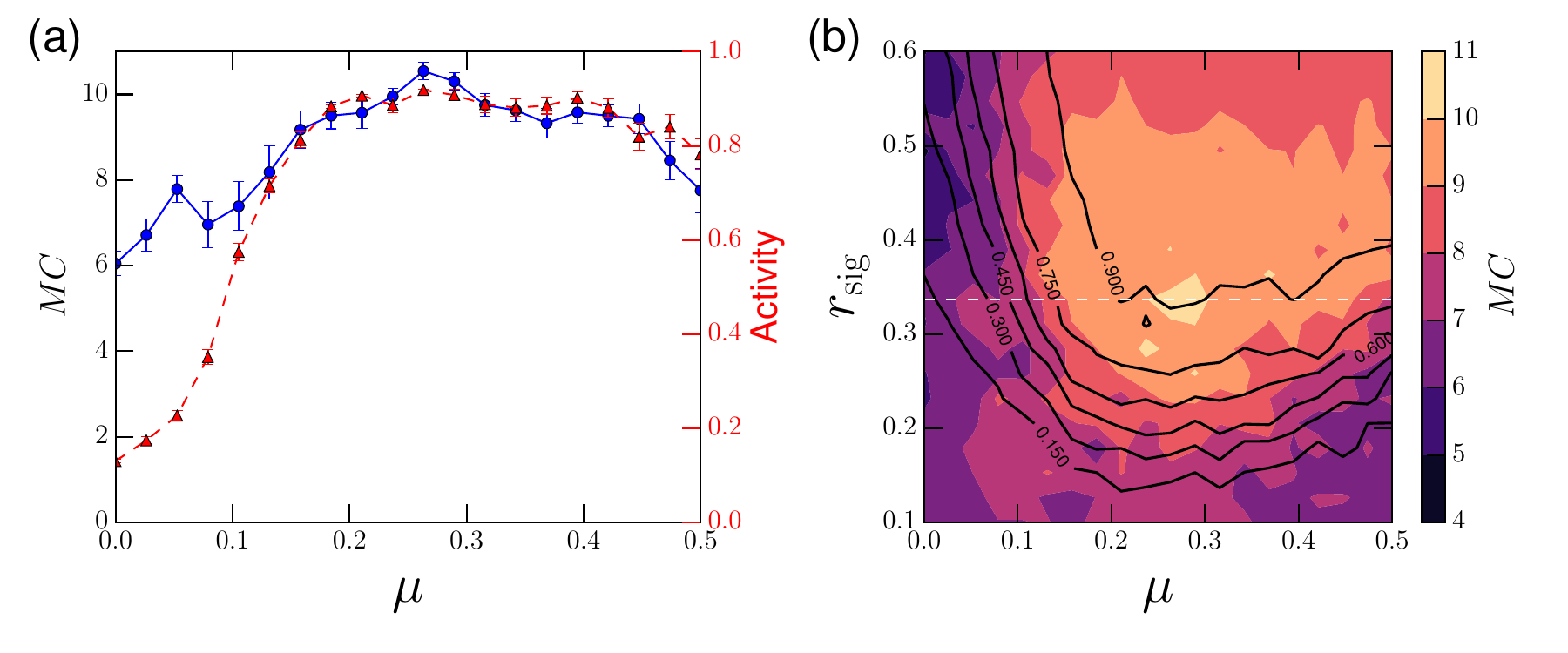}
\caption{To verify that optimal $\mu$ for both performance and diffusion coincide, we overlay the contours of the MC task and the information diffusion process. (a) A slice through the contour at $r_{\textup{sig}}=0.3$ shows that the normalized activity is maximized when $MC$ is maximized. (b) The solid black lines show the contours for the information diffusion process, closely matching the $MC$ contours for performance on the task. In this information diffusion task an input identical to the one given the MC task is used so that $r_{\textup{sig}}$ correspond to the same levels of activation the network would receive during the MC task. The total sum of activation is then used, normalized by the duration of the test and the size of the network. Error bars represent the standard error of the mean.}
\label{fig:supp_overlap}
\end{figure}

\begin{figure}[tbhp!]
\centering
\includegraphics[width=0.5\linewidth]{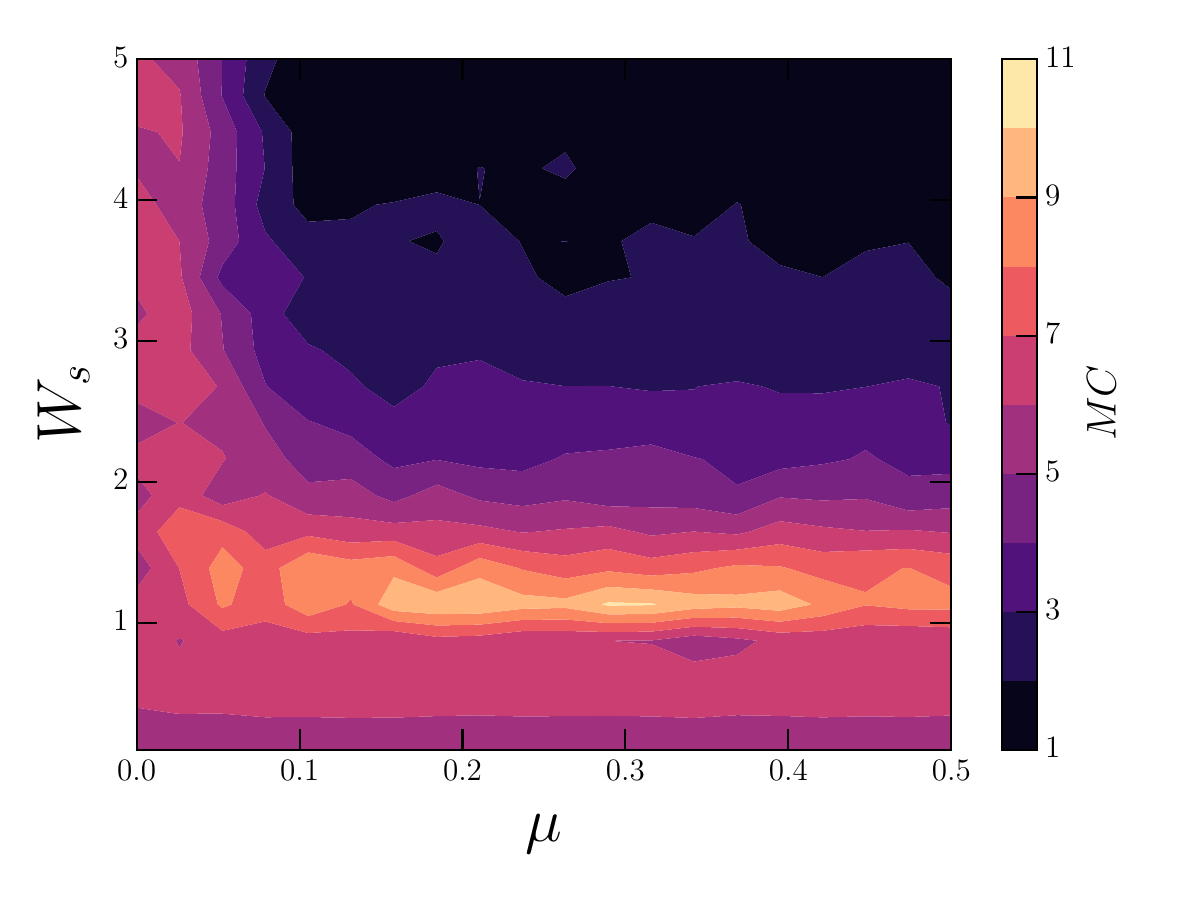}
\caption{A contour of the $MC$ performance on the same network as in the main-text for the MC task while holding $r_{\textup{sig}}=0.3$ and varying over the parameter $W_s$, a constant factor that the weight matrix $\boldsymbol{W}$ is multiplied by. Best performance occurs within a narrow band of weights. Importantly, there are no weights either above or below the optimal $\mu$ region where performance reaches the same level, suggesting that controlling for the size of the largest eigenvalue of the adjacency matrix will be insufficient to explain the performance gains afforded by the modularity.}
\label{fig:supp_ws}
\end{figure}

\begin{figure}[tbhp!]
\centering
\includegraphics[width=0.5\linewidth]{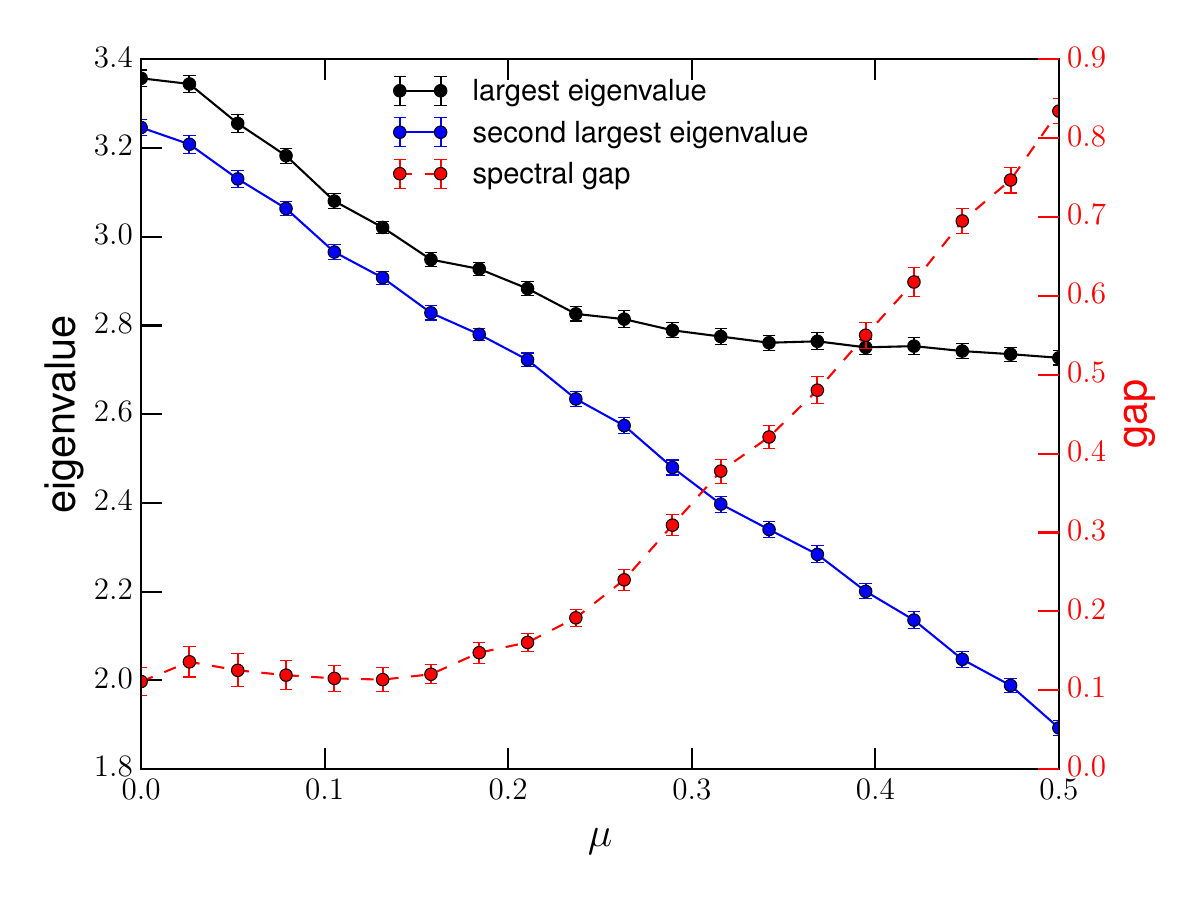}
\caption{The absolute values of the largest and second largest eigenvalues of the adjacency matrix and spectral gap are plotted for the same network used in the MC task. At low $\mu$ the gap is small and both the first and second eigenvalues decrease at roughly the same rate as $\mu$ increases. However, at roughly $\mu \approx 0.25$ the rate of decline of the largest eigenvalue flattens out and the gap increases as the network approaches a traditional random graph. Error bars represent the standard error of the mean.}
\label{fig:supp_spectrum}
\end{figure}

\begin{figure}[tbhp!]
\centering
\includegraphics[width=0.5\linewidth]{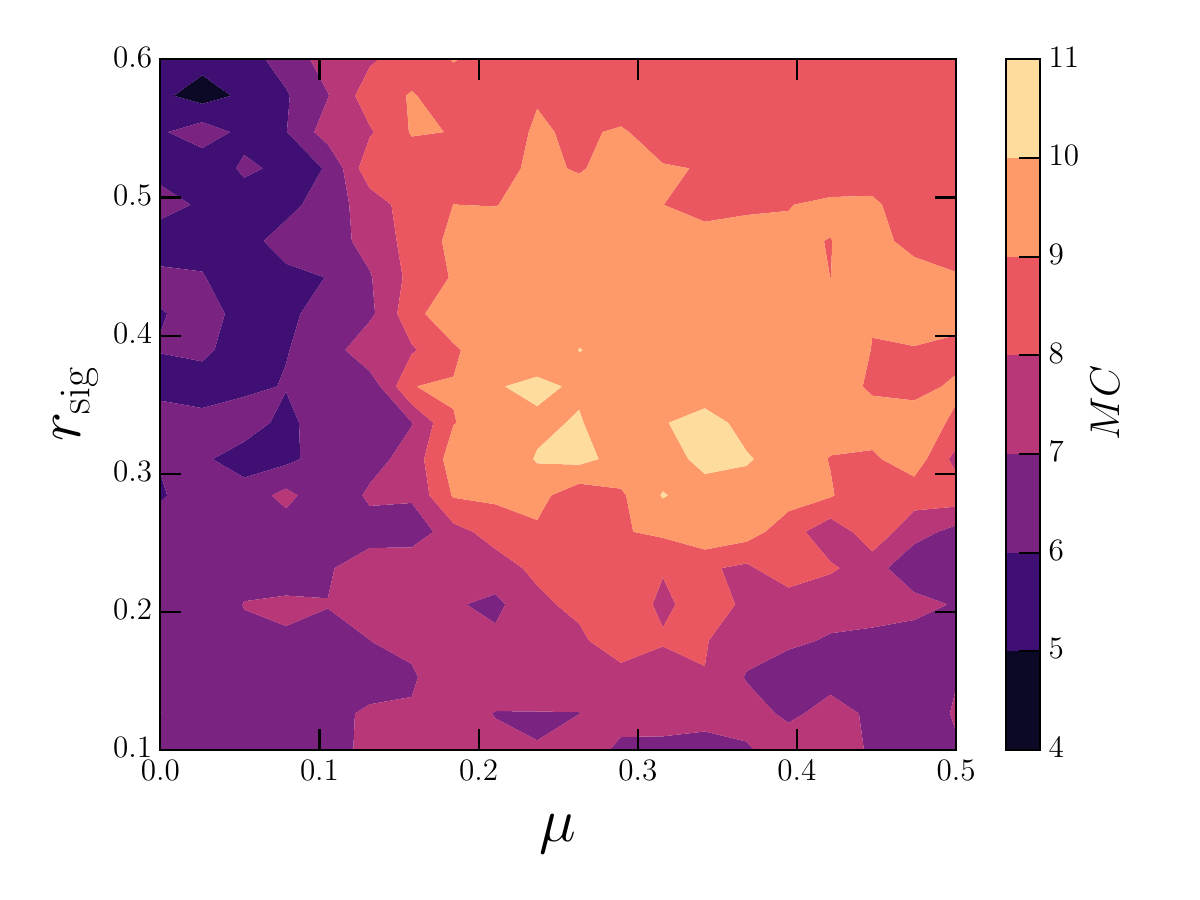}
\caption{We can correct for the changes in the spectral radius as $\mu$ is changed by holding the absolute value of the largest eigenvalue constant. Qualitatively we get the same results even while maintaining a constant spectral radius. The inability of the spectral radius to account for the increased memory capacity is likely due to our use of threshold-like activation functions that mimic complex contagions. Traditionally, the strong relationship between spectral radius and memory capacity is linked with traditional tanh activation functions.}
\label{fig:supp_spectral_rsig}
\end{figure}

\begin{figure}[tbhp!]
\centering
\includegraphics[width=0.9\linewidth]{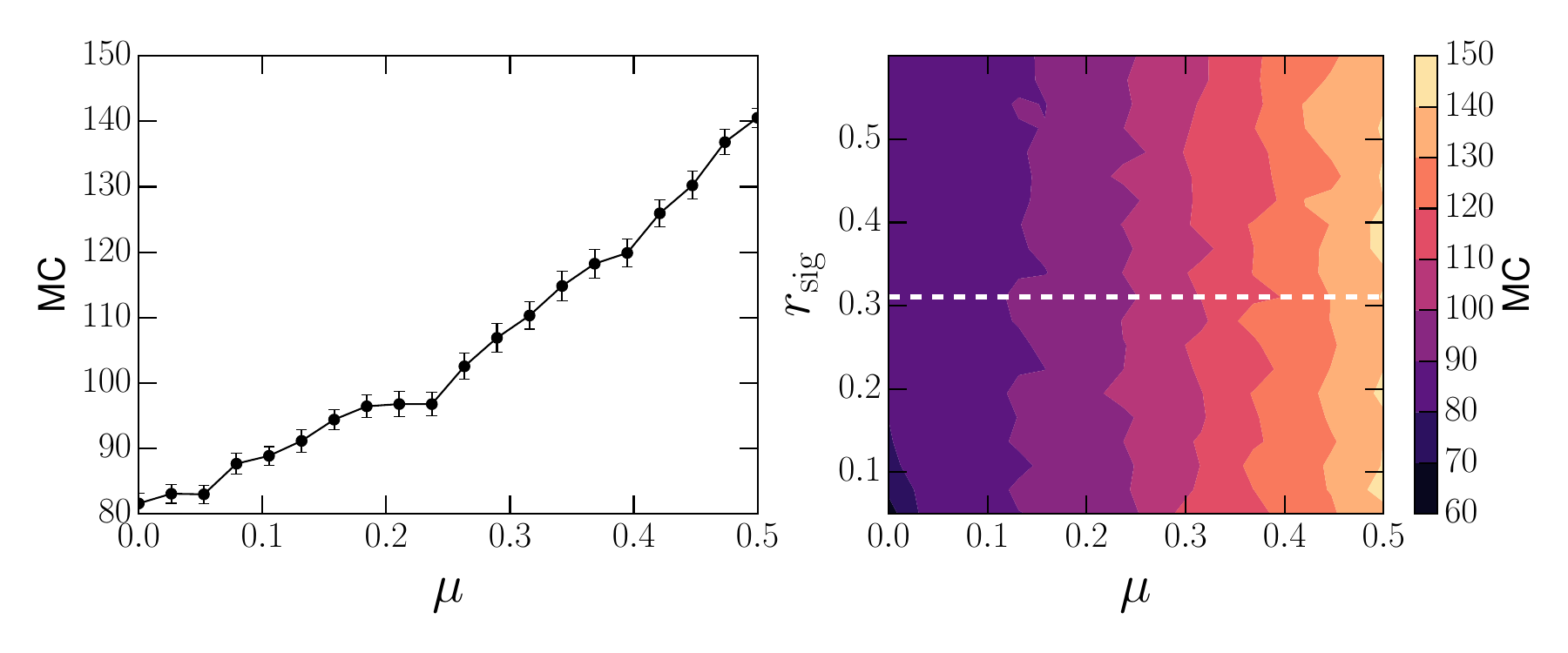}
\caption{The memory capacity ($MC$) of a linear reservoir. The introduction of community structure degrades memory performance. (left) Shows a slice through the phase diagram at the dashed white line. (right) The memory capacity as $\mu$ and input fraction vary. The same graphs and parameters as the activation simulation are used; with $N=500$, $k=6$, a community size of $10$, a $W_s=0.9$, and with corrections for the spectral radius.}
\label{fig:linear_mc_results}
\end{figure}

\begin{figure}[tbhp!]
\centering
\includegraphics[width=0.9\linewidth]{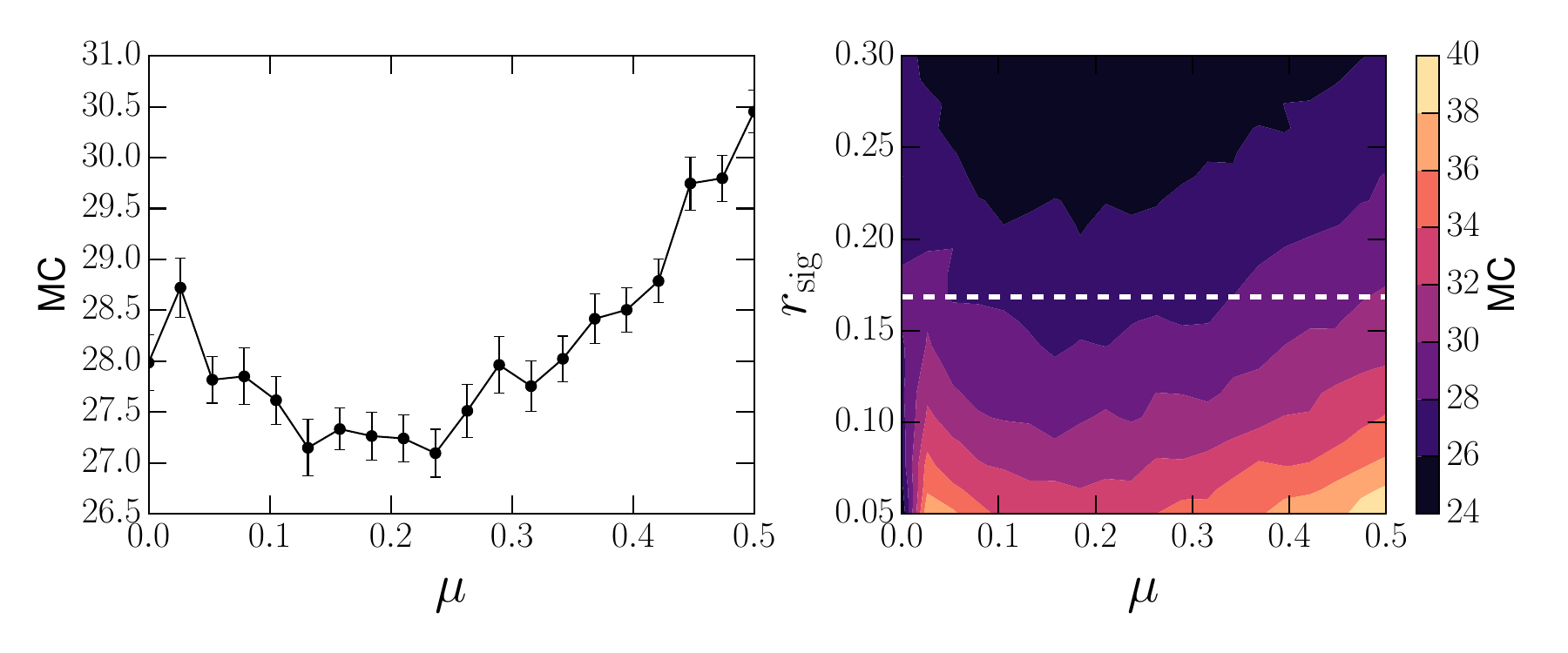}
\caption{The memory capacity ($MC$) of a tanh reservoir. Similar to the linear reservoir, the introduction of community structure degrades memory performance. As $r_\textup{sig}$ decreases, the memory capacity rises due to the reduced external excitations allowing the reservoir to operate closer to the higher performing linear regime of the tanh activation function. (left) Shows a slice through the phase diagram at the dashed white line. (right) The memory capacity as $\mu$ and input fraction vary. The same graphs and parameters as the activation simulation are used; with $N=500$, $k=6$, a community size of $10$, a $W_s=1.0$, and with corrections for the spectral radius.}
\label{fig:tanh_mc_results}
\end{figure}

\end{document}